\documentclass{article} % For LaTeX2e
\usepackage{iclr2027_conference,times}

\usepackage{amsmath,amsfonts,bm}

\def\eqref#1{equation~\ref{#1}}
\def\1{\bm{1}}

\DeclareMathAlphabet{\mathsfit}{\encodingdefault}{\sfdefault}{m}{sl}
\SetMathAlphabet{\mathsfit}{bold}{\encodingdefault}{\sfdefault}{bx}{n}

\usepackage{hyperref}
\usepackage{url}

\usepackage{cleveref}

\usepackage{bm}
\usepackage{amsmath}
\usepackage{amssymb}
\usepackage{amsthm}

\usepackage{multirow}
\usepackage{multicol}
\usepackage{makecell}
\usepackage{wrapfig}

\usepackage{adjustbox}
\usepackage{booktabs}
\usepackage{threeparttable}
\usepackage{tabularx}
\usepackage{subcaption}
\usepackage[table]{xcolor}

\usepackage{algorithm}
\usepackage{algorithmicx}
\usepackage[noend]{algpseudocode}
\usepackage{bbding}

\usepackage{pifont}
\usepackage{mathtools}
\usepackage{thmtools}
\usepackage{thm-restate}

\usepackage{titlesec}
\usepackage{enumitem}
\usepackage{graphicx}

\definecolor{algblue}{RGB}{65,125,170}
\definecolor{algred}{RGB}{215,95,95}
\definecolor{algorange}{RGB}{195,130,65}
\definecolor{algteal}{RGB}{65,145,135}
\definecolor{algpurple}{RGB}{145,110,175}

\newcommand{\samplername}{HA-Sampler}
\newcommand{\rewardname}{HA-Scorer}

\renewcommand{\paragraph}[1]{\vspace{.2em}\noindent\textbf{#1.}}

\titlespacing*{\section}
  {0pt}      % 左缩进
  {0.2em}    % section 标题前间距
  {0.2em}    % section 标题后间距
\titlespacing*{\subsection}{0pt}{0.2em}{0.1em}
\titlespacing*{\subsubsection}{0pt}{0.2em}{0.1em}

\title{Risk Decomposition for Hallucination-Aware Video Sampling}

\author{
\normalfont
\makebox[\dimexpr\textwidth-2\tabcolsep\relax][c]{%
\begin{tabular}{@{}l@{}}
\resizebox{\dimexpr\textwidth-2\tabcolsep\relax}{!}{%
\small\bfseries
Chenyan Zheng\textsuperscript{1,*}\quad
Fan Wu\textsuperscript{2,*,\ensuremath{\dagger}}\quad
Ruibo Li\textsuperscript{2}\quad
Jiaxing Li\textsuperscript{2}\quad
Minghao Yang\textsuperscript{3}\quad
Bang Zhang\textsuperscript{4}\quad
Guosheng Lin\textsuperscript{2,\ensuremath{\ddagger}}
}\\[0.3em]
\multicolumn{1}{c}{
\footnotesize
\textsuperscript{1}Zhejiang University\quad
\textsuperscript{2}Nanyang Technological University\quad
}\\[0.3em]
\multicolumn{1}{c}{
\footnotesize
\textsuperscript{3}Tsinghua University\quad
\textsuperscript{4}Tongyi Lab, Alibaba Group\quad
}\\[0.3em]
\multicolumn{1}{c}{
\footnotesize
\textsuperscript{*}Equal contribution\quad
\textsuperscript{\ensuremath{\dagger}}Project lead\quad
\textsuperscript{\ensuremath{\ddagger}}Corresponding author
}\\[0.3em]
\multicolumn{1}{c}{
\small\textcolor{magenta}{\texttt{https://github.com/wufan-cse/HA-Sampling}}
}
\end{tabular}
}
}

\iclrfinalcopy % Uncomment for camera-ready version, but NOT for submission.
\begin{document}

\maketitle
\lhead{Preprint.} % 清空顶部的会议声明

\begin{abstract}
Despite achieving high visual fidelity, video models still suffer from physical hallucinations, limiting their reliability as world simulators.
Existing methods leverage model-internal or external signals to improve physical plausibility, but their roles and limitations require a unified explanation.
We propose a two-dimensional physical-hallucination risk framework that defines low-density sampling risk and high-density drifting risk.
The first risk captures generations assigned low likelihood by the video model, for example, a ball suddenly teleporting despite the model favoring continuous motion, while the second risk captures physical inconsistencies in high-likelihood generations, for example, a ball
bouncing progressively higher without additional energy.
Our analysis explains how model-internal signals steer sampling toward higher-density regions of the learned distribution to address low-density sampling risk.
However, higher model likelihood does not guarantee physical plausibility, motivating external guidance to address high-density drifting risk.
Based on this framework, we develop Hallucination-Aware Sampler (\samplername{}), which applies external physical guidance within latent refinement to generate physically plausible videos, and \rewardname{}, which derives a likelihood-based signal from the generator itself for evaluating generated videos using continuous normalizing flow (CNF), with coarse ODE integration and finite-difference divergence estimates.
Experiments on Physics-IQ and VideoPhy benchmarks support our theoretical analysis and demonstrate that our methods outperform the compared sampling and scoring SOTA baselines.
% in generating and selecting physically realistic videos.
% Experiments on Physics-IQ and VideoPhy benchmarks indicate the effectiveness of our proposed theory and the state-of-the-art performance of our proposed methods demonstrate the effectiveness on generating and selecting physically real videos.
\end{abstract}

\begin{figure}[h]
    \centering
    \includegraphics[width=\linewidth]{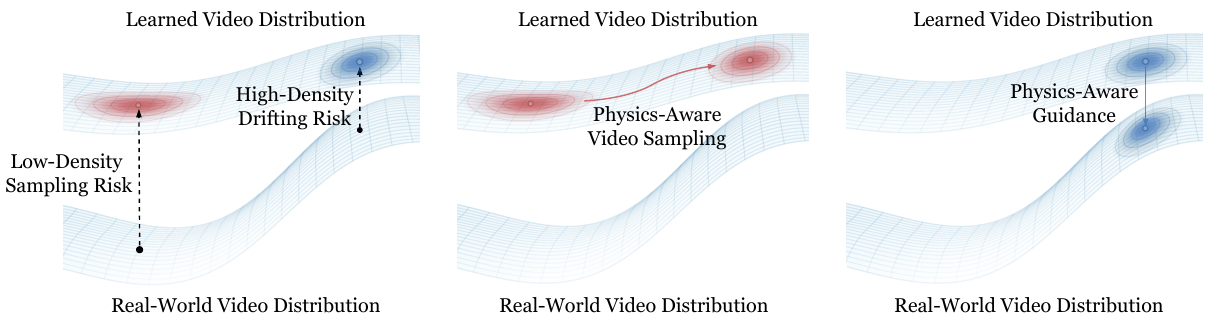}
    \caption{\textbf{Overview of our two-dimensional physics hallucination risk framework.} \textbf{Left:} Low-density sampling risk captures generations assigned low likelihood by the video model, while high-density drifting risk captures physical errors in high-likelihood generations.
    \textbf{Middle:} Model-internal physics-aware signals steer sampling toward higher-density regions of the learned distribution to reduce low-density sampling risk.
    \textbf{Right:} External physics-aware guidance helps correct the drifting between model-preferred generation and physical plausibility to address high-density drifting risk.}
    \label{fig:overview}
\end{figure}

\section{Introduction}
\label{sec:introduction}

% \makeatletter
% \typeout{CHECK-FONT-SIZE: \f@size pt}
% \typeout{CHECK-BASELINE: \the\baselineskip}
% \makeatother
%%%%%%%%%%%%%%% 09.20 %%%%%%%%%%%%%%%
% Contribution: 我们提出统一的视频物理幻觉风险框架，说明模型内部信号与外部信号的作用机制及其局限，解释了二者在缓解幻觉中的互补关系

% 现有工作认为模型内部存在信号对帮助提升生成真实性有正向作用，同时利用外部信号也能够帮助生成效果。
% 本文我们提出了统一的理论解释框架，首次从理论上定义了这种内部信号并解释了内外两种信号的互补作用机制，相比前人工作，我们证明这种内部信号对提高物理真实性并不是一直有正向作用，在模型高度支持的物理错误会失效，外部信号能够弥补内部信号的失效。

% 因此takeaway就是：内部信号能够帮助改善模型自身不充分支持的生成（第一类），但不能排除模型高度支持的物理错误（第二类）；有效的幻觉缓解需要利用前者，同时借助外部证据检验后者。

% 三个价值点：
% 1. 对后续方法研发，如果自修正不断提高模型似然，却不再改善独立评估的物理正确性，就提示瓶颈可能不在采样，而在模型对物理规律的认识，没学好，继续强化同一先验未必有效，应考虑外部约束或针对性训练。
% 2. 对生成和reward，模型能提供候选视频质量的内部信号判断，可以与外部验证共同用于修正和筛选
% 3. 对下游应用，例如机器人规划或合成训练数据，可以进一步研究用内外部两种信号识别需要复核的生成结果。
%%%%%%%%%%%%%%% 09.20 %%%%%%%%%%%%%%%

% Recent video models have made substantial progress in visual fidelity and motion generation~\TODO{cite} as world simulators for generating physical realism training data.
% 在robotics这类追求高精度的下游应用有重要的作用。
% However, physical hallucinations---such as disappearing objects and impossible interactions---undermine their reliability in these applications~\TODO{cite}.
% 现有方法xxx做了很多努力解决这个问题，主要分为两类，利用模型内部存在信号对帮助提升生成真实性有正向作用，同时利用外部信号也能够帮助生成效果。
Recent advances in visual fidelity and motion generation have highlighted the potential of video models as world simulators for downstream tasks, such as robotics~\citep{yang2026rise}.
Such applications require accurate physical plausibility, yet physical hallucinations---such as disappearing objects and impossible interactions---undermine the reliability of generated videos~\citep{bansal2025videophy,motamed2026physicsiq}.
To address this, existing methods for video sampling can be categorized into two directions: one~\citep{hassan2026proprio,yuan2026likephys} suggests that video models' own preferences contain useful information about physical plausibility and leverage model-internal signals~\citep{jang2026selfrefining, cho2026tag} for refining, while the other uses external signals~\citep{yuan2026wmreward,interno2026geophys} to guide generation and filter physically unreal samples.
% 现有工作认为模型内部存在信号对帮助提升生成真实性有正向作用，同时利用外部信号也能够帮助生成效果。

% 讲两类risk
% However, 现有工作并没有系统性分析这些信号的作用机制。
% In this work, we introduce a unified framework for understanding这些作用机制。
% 我们定义了low-density和high density来衡量发生physical hallucinations的风险，as shown in fig1 left.
% 第一类low-density指的是真实世界分布中存在的少量情况但物理真实的例子，另一种说法是distribution long tail问题，生成这类数据的时候模型存在发生幻觉的风险因为没有充分学习。
% 第二类high density相对于low density来说，模型在生成自己充分学习过的数据pattern时并没有严格遵循这类数据pattern，发生了漂移（这里参考一下模式插值这篇文章是怎么讲的）。
However, these raise a fundamental question: how do model-internal and external signals address physical hallucinations?
We propose a unified framework that defines two-dimensional risks of physical hallucination to study and understand this question, as shown in~\Cref{fig:overview} (left).
Low-density sampling risk captures the tendency to generate samples with low likelihood under the video model.
For example, a model that favors continuous ball motion may nevertheless generate a low-likelihood trajectory in which the ball suddenly jumps from one location to another.
On the contrary, high-density drifting risk captures physical inconsistencies in samples assigned high likelihood by the video model.
For example, the model may favor a video of a ball bouncing progressively higher on a floor, despite no additional energy.
The former reflects a deviation from the model's own preferences,
whereas the latter exposes a drifting between those preferences and physical plausibility.

% 在我们这套框架下，我们分析了model-internal signal和external signal的作用机制，发现model internal signal解决的是low density risk而external signal则作用在high density risk上。
% model-internal signal的作用方式是xxx
% model-external signal的作用方式是xxx
% Compared with existing methods, 我们还额外发现了model internal signal并非一直有效，当幻觉来源dominated by第二类risk的时候，model internal signal会失效，换一个说法，high level解释。
% meanwhile，当幻觉来源dominated by第一类risk的时候，model-external signal也会失效，从另一个角度说，对于模型没充分学习的sample无论怎么guide都无法让模型生成physically real的sample。
In our framework, we analyze how model-internal and external signals address the two risk dimensions, respectively.
Model-internal signals steer video sampling toward the model's own preference region, which is higher-density regions of the learned video distribution, targeting the low-density sampling risk, as shown in~\Cref{fig:overview} (middle).
However, these regions may drift from physical plausibility.
% reinforcing the model's preferences can preserve physical errors that it already assigns high likelihood.
External signal provides a physical guidance, forcing generations toward physically plausible videos from the model's own preference, addressing high-density drifting risk, as shown in~\Cref{fig:overview} (right).
Our theory reveals the improvement by either signal alone is not sufficient.

% 在本文中，我们提出了一套系统性框架来understanding和分析这个问题，先介绍各种名词以及对问题等的定义。
% 然后讲解我们的理论系统

% 基于我们这套框架，对于现有方法我们有一个全新的结论，证明这种内部信号对提高物理真实性并不是一直有正向作用，在模型高度支持的物理错误会失效，外部信号能够弥补内部信号的失效。
Furthermore, we propose a hallucination-aware video sampler (\samplername), and scorer (\rewardname).
\samplername~generates physically plausible samples by integrating external physical guidance into model-based refinement, targeting both dimensions of hallucination risk.
% \rewardname~estimates the generator's conditional likelihood using sparse continuous normalizing flow (sparse-CNF), with coarse ordinary differential equation (ODE) integration and finite-difference divergence estimates, resulting model's own preferences candidates.
\rewardname~derives a likelihood-based signal from the generator itself for evaluating generated videos, using continuous normalizing flow (CNF), with coarse ordinary differential equation (ODE) integration and finite-difference divergence estimates.
Our extensive experiments show that~\samplername~and~\rewardname~surpass state-of-the-art (SOTA) baselines, improving the physical plausibility of generated videos and helping select more physically plausible candidates, respectively.
% Furthermore，基于我们的理论框架以及此结论的启发，我们做了两类方法，sampling和reward。介绍具体的做法。
% 我们设计了一系列实验，实验结果证明了我们理论分析的正确性以及两种方法的有效性，sota
Our contributions can be summarized as follows:
\begin{enumerate}
[
    itemsep=2pt,     % 不同 \item 之间的额外间距
    topsep=1pt,      % 列表与上下正文之间的额外间距
    parsep=0pt,      % 同一 \item 内不同段落之间的间距
    partopsep=0pt    % 列表另起一段时增加的额外间距
]
    \item We are the first to introduce a unified physical-hallucination risk framework that defines low-density sampling risk and high-density drifting risk.
    \item We identify and demonstrate that video model-internal signals contain useful information related to physical plausibility that helps mitigate the first type of risk, but these signals are not always effective. 
    External guidance can complement internal signals when they fall short by solving the second type of risk.
    % 模型内部存在physical realism相关的信号并且这种内部信号作用于risk1，但这种信号并非一直有效。另一方面，外部信号guidance作用于risk2，该信号可在内部信号失效时作为补充
    \item Based on our framework, we propose a hallucination-aware sampler,~\samplername, and scorer,~\rewardname, for generating and scoring physically plausible videos, respectively. 
    Our experiments show that these two proposed methods achieve the SOTA performance. 
\end{enumerate}

% 文章逻辑
% 1. 我们提出hallucination risk来衡量视频生成模型发生幻觉的风险
% 2. hallucination risk由两种condition组成，type A和type B
% 3. 现有的方法分为两类，第一类方法只能解决type A，如pnp，第二类方法只能解决type B，如wmreward，没有一个方法能同时解决这两种types
% 我们的contribution：
% 1. 定义了一套新的统一的理论框架，兼容并扩充了之前的理论体系
% 2. 新定义的hallucination risk可以作为一种指导性的指标，一是对比不同的模型(应该是sampler?)的幻觉程度。二是一种训练观察指标。
% 3. 基于这套理论，我们设计了一种能同时解决两种types的方法
% 

% Recent video generative models~\cite{wan2025, kong2024hunyuanvideo, hacohen2025ltxvideo, gao2025seedance, kondratyuk2024videopoet} have achieved remarkable progress in visual fidelity and temporal coherence, enabling realistic generation from text or image conditions. 
% % However,  violate basic physical dynamics, such as object gravity and fluid-object interaction. 
% However, these models still suffer from hallucination~\cite{mei2026videogenerationmodelsrobotics, motamed2026physicsiq, bansal2026videophy2}, where the generated video appears visually plausible but deviates from physical realism.
% These failures limit the use of video generative models for frontier applications, such as world model~\cite{nvidia2025cosmos} and embodied AI~\cite{yang2026rise}.
% embodied AI, robot planning, and physical reasoning, where generated futures must be not only realistic-looking but also physically plausible.

%% overview

\section{Related Work}
\label{sec:related_work}

\subsection{Video generative models}
Recent video generative models have achieved substantial progress in visual fidelity, temporal coherence, and controllability~\citep{wan2025,blattmann2023stable,chen2025skyreelsv2,hacohen2025ltxvideo}. 
They are commonly studied under two settings: text-to-video (T2V)~\citep{chen2023videocrafter1,singer2022make,hong2022cogvideo,tian2024videotetris}, and image-to-video (I2V)~\citep{blattmann2023stable,wan2025,hu2022make,yang2025cogvideox,hacohen2025ltxvideo,wu2026ic}. 
% Compared with T2V, I2V provides stronger visual grounding by fixing the initial scene state, making it particularly relevant to world-model and embodied applications~\cite{mei2026videogenerationmodelsrobotics}. 
However, both T2V and I2V models still suffer from hallucination~\citep{motamed2026physicsiq,li2025worldmodelbench,lin2025exploring,kang2024far}: generated videos are hard to preserve physical plausibility.

\subsection{Physics-Aware Video Generation}

% Our paper explicitly distinguishes low-density and high-density risks, characterizing the roles and limitations of model-internal and external signals in generating physically real videos.

\paragraph{Model-internal guidance}
Methods based on model-internal guidance~\citep{bai2025zigzag,wu2024freeinit,jang2026selfrefining, shaulov2025flowmo} aim to enable models to refine themselves without external signal or verification.
% In image diffusion models, Zigzag Diffusion Sampling~\cite{bai2025zigzag} introduces a self-reflection mechanism by alternating denoising and inversion steps during sampling to accumulate prompt-related semantic information.
For video diffusion models, FreeInit~\citep{wu2024freeinit} refines the initial noise through an iterative denoise and re-noise procedure, which improves temporal consistency without training.
Self-Refining Video Sampling~\citep{jang2026selfrefining} interprets flow-matching video generators as denoising autoencoders and proposes Predict-and-Perturb (P\&P) steering the sampling trajectory toward more stable and self-consistent modes of the pretrained model.
Tangential Amplifying Guidance (TAG)~\citep{cho2026tag} amplifies the tangential component of each sampling update while preserving its radial component, steering samples toward regions of higher model density.
Zigzag Diffusion Sampling~\citep{bai2025zigzag} proposes a sampling strategy that alternates denoising and inversion steps, allowing pretrained diffusion models to accumulate semantic information. 
FlowMo~\citep{shaulov2025flowmo} improves motion coherence by extracting temporal representations from the model's own predictions and reducing patch-wise temporal variance during sampling.
However, reinforcing the model's own preferences does not guarantee physical plausibility, as physically incorrect outcomes may still suffer from high model likelihood.

\paragraph{External guidance}
% Recent efforts on using external guidance to generate physically real videos can be categorized into three groups. 
% One line of work focuses on injecting explicit physical priors or simulation-grounded supervision into generation~\citep{liu2024physgen,xiong2026physalign,lu2026phys4d}.
% PhysGen~\citep{liu2024physgen} combines rigid-body simulation with image-to-video generation while PhysAlign~\citep{xiong2026physalign} leverages synthetic data with physical and 3D annotations to align visual features with physical representations.
% The second line of work~\citep{zhang2026physrvg,cai2025phygdpo} improves physical consistency through training-time reward optimization, including PhysRVG~\citep{zhang2026physrvg} which introduces a physics-aware reinforcement learning framework that enforces collision-related physical rules, and PhyGDPO~\citep{cai2025phygdpo}
% applies groupwise preference optimization with VLM-based physics rewards.
% Phys4D~\citep{lu2026phys4d} extends video diffusion models to physics-consistent 4D representations.
% The third line of work~\citep{zhang2025videorepa,yuan2026wmreward} transfers physics knowledge from pretrained representation models. 
% WMReward~\citep{yuan2026wmreward} leverages a latent world model as an inference-time reward for physics alignment.
Recent efforts on using external guidance introduce physical knowledge through additional supervision, pretrained models, or reward functions.
PhysAlign~\citep{xiong2026physalign} uses simulation-derived annotations and video-feature alignment, while GeoPhys~\citep{interno2026geophys} assesses physical plausibility through the geometry of temporal feature trajectories from frozen image encoders.
% while Phys4D~\citep{lu2026phys4d} develops physics-consistent 4D representations through pseudo-supervised pretraining, simulation-grounded fine-tuning, and reinforcement learning.
PhysRVG~\citep{zhang2026physrvg} and PhyGDPO~\citep{cai2025phygdpo} improve physical consistency through physics-aware reinforcement learning and VLM-based preference optimization, respectively.
VideoREPA~\citep{zhang2025videorepa} transfers physical knowledge from video foundation models through relational alignment.
At inference time, WMReward~\citep{yuan2026wmreward} uses a latent world model to score candidates and guide denoising.
However, improving an external physical score does not necessarily increase model likelihood or reduce low-density sampling risk.

% However, these methods often depend on additional training or simulation data, making them computationally expensive and difficult to transfer across different scenarios.

% However, most existing methods are often tied to domain-specific physical priors or provide limited improvements in physical consistency.
% World model and embodied applications require generated videos to be physically aligned with real-world dynamics, yet current video generative models still struggle to faithfully model physical laws.
% VideoREPA~\cite{zhang2025videorepa} distills physics-related token relations from video understanding foundation models.

\paragraph{Hybrid strategies}
Hybrid strategies coordinate model-internal signal with external guidance.
% Universal Guidance~\citep{bansal2024universal} introduces per-step self-recurrence to improve sample fidelity under external guidance.
% FreeDoM~\citep{yu2023freedom} couples energy-based guidance with time-travel resampling to improve conditional generation.
Trust Sampling~\citep{huang2024constrained} monitors the model's predicted noise magnitude to terminate external optimization when samples depart from its estimated state manifold.
TFG~\citep{ye2024tfg} further unifies guidance and recurrence within a common algorithmic design space.
However, beyond coordinating these mechanisms, our framework explicitly distinguishes low-density sampling risk from high-density drifting risk, explaining the complementary roles and limitations of model-internal and external signals in mitigating physical hallucinations.

% Our paper explicitly distinguishes low-density and high-density risks, characterizing the roles and limitations of model-internal and external signals in generating physically real videos.

% However, jointly using internal and external signals does not by itself explain which sources of physical hallucination each addresses or where either falls short.
% Our framework distinguishes low-density sampling risk from high-density drifting risk, providing a theoretical account of their complementary roles and limitations.

% Our framework provides a risk-based interpretation of this
% coordination, distinguishing low-density sampling risk from
% high-density drifting risk and characterizing the complementary
% roles and limitations of the two signals.

% \clearpage

\section{Methodology}
% 注释：endpoint指的是一个noisy latent对应的去噪轨迹的重点
\subsection{Notations}

% Let $\mathbf{x} \in \mathcal{X}$ be a clean video under condition $\mathbf{c}$ and $\mathbf{z}_0 = \mathcal{E}(\mathbf{x}) \in \mathcal{Z}$ its latent, where $\mathcal{E}$ and $\mathcal{G}$ are the frozen VAE encoder and decoder.
% Conditioned on $\mathbf{c}$, $q^\star$ and $q$ denote the real-video and training latent distributions, while $p_\theta$ and $p_{\theta,\mathcal{S}}$ denote the model's output distributions under ideal sampling and a practical sampler $\mathcal{S}$, respectively.

% For $t \in [0,1]$, we adopt the data-to-noise flow matching path
% \begin{equation}
% \mathbf{Z}_t = (1-t)\mathbf{Z}_0 + t\mathbf{Z}_1, \qquad \mathbf{Z}_0 \sim q(\cdot \mid \mathbf{c}), \quad \mathbf{Z}_1 \sim \mathcal{N}(\mathbf{0},\mathbf{I}),
% \label{eq:fm_path}
% \end{equation}
% with target velocity $\mathbf{V} = \mathbf{Z}_1 - \mathbf{Z}_0$.
% Generation integrates the learned velocity field $v_\theta(\mathbf{z}_t,t,\mathbf{c})$ from $t=1$ to $t=0$, with clean endpoint prediction
% \begin{equation}
% \widehat{\mathbf{z}}_{0\mid t} = \mathbf{z}_t - t\,v_\theta(\mathbf{z}_t,t,\mathbf{c}).
% \label{eq:clean_predictor}
% \end{equation}

Let $\mathbf{z}_0 = \mathcal{E}(\mathbf{x}) \in \mathcal{Z}$ be the latent of a clean video $\mathbf{x} \in \mathcal{X}$, with frozen VAE encoder $\mathcal{E}$ and decoder $\mathcal{G}$.
Conditioned on $\mathbf{c}$, $q^\star$ and $q$ denote real-video and training latent distributions, while $p_\theta$ and $p_{\theta,\mathcal{S}}$ denote model distributions under ideal sampling and a practical sampler $\mathcal{S}$.
We use the data-to-noise flow-matching path $\mathbf{Z}_t = (1-t)\mathbf{Z}_0 + t\mathbf{Z}_1$, $t \in [0,1]$, with $\mathbf{Z}_0 \sim q(\cdot \mid \mathbf{c})$, $\mathbf{Z}_1 \sim \mathcal{N}(\mathbf{0},\mathbf{I})$, and target velocity $\mathbf{V} = \mathbf{Z}_1 - \mathbf{Z}_0$.
Generation integrates the learned velocity field $v_\theta(\mathbf{z}_t,t,\mathbf{c})$ from $t=1$ to $t=0$, with clean endpoint prediction $\hat{\mathbf{z}}_{0\mid t} = \mathbf{z}_t - t\,v_\theta(\mathbf{z}_t,t,\mathbf{c})$.
% \begin{equation}
% \hat{\mathbf{z}}_{0\mid t} = \mathbf{z}_t - t\,v_\theta(\mathbf{z}_t,t,\mathbf{c}).
% \label{eq:clean_predictor}
% \end{equation}

%% 这句话可能不太需要
% We denote the physical-inconsistency score of a world model by $S_{\phi}(\mathcal{G}(\mathbf{z}),\mathbf{c})$, where a lower score indicates greater physical plausibility.

\subsection{Preliminaries}
% \paragraph{Flow-matching video generation.}
% Flow-matching has become an efficient alternative to diffusion-based approaches for video generation\TODO{cite}.
% This paradigm transforms the Gaussian noise distribution into the data distribution in the latent space.
% For a latent representation $\mathbf{z}_t \in \mathbb{R}^{F\times H\times W\times C}$,
% where \(F\), \(H\), \(W\), and \(C\) denote the number of latent frames, height, width, and channels, respectively, 
% flow-matching learns a velocity field $v_\theta(\mathbf{z}_t, t, \mathbf{c})$ that defines an ordinary differential equation (ODE):
% \begin{equation}
%     \frac{d\mathbf{z}_t}{dt} = v_\theta(\mathbf{z}_t, t, \mathbf{c}),
% \end{equation}
% The model is trained to approximate this target velocity:
% \begin{equation}
%     \mathcal{L}_{\mathrm{FM}}(\theta)
%     =
%     \mathbb{E}_{\mathbf{z}_0,\mathbf{z}_1,t,c}
%     \left[
%     \left\|
%     v_\theta(\mathbf{z}_t,t,\mathbf{c}) - (\mathbf{z}_1 - \mathbf{z}_0)
%     \right\|_2^2
%     \right].
% \end{equation}

\paragraph{Mode interpolation}
% \paragraph{Hallucination set and mode interpolation.}
In prior work~\citep{aithal2024understanding}, the $\epsilon$-hallucination set is defined below.
\begin{equation}
    \label{hallucination_set}
        H_\epsilon(q^*) = \{ \mathbf{z}_0 : q^*(\mathbf{z}_0 \mid \mathbf{c}) \leq \epsilon \}.
\end{equation}
The parameter $\epsilon$ is taken to be vanishingly small. That is, if $\mathbf{z}_0\in H_\epsilon(q)$, it lies outside of the support of the real data distribution, 
which means that the generated video was a hallucination case. The complement of the hallucination set is the $\epsilon$-support set denoted by $S_\epsilon(q^*)$.
Mode interpolation is a common phenomenon in diffusion models.
This phenomenon occurs when the model generates samples that are interpolations of multiple modes in the $\epsilon$-support set, such that the generated sample lies in the $\epsilon$-hallucination set. 
For example, if $x, y\in S_\epsilon(q^*)$, then the model generates a sample $z = \alpha x + (1-\alpha) y$ for some $\alpha\in(0,1)$, and $z\in H_\epsilon(q^*)$. 

\begin{figure}[t]
    \centering
    \begin{subfigure}[b]{\linewidth}
        \centering
        \includegraphics[
            width=\linewidth,
        ]{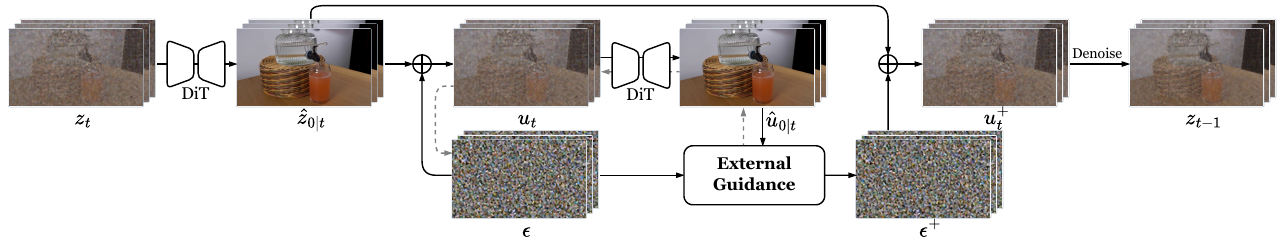}
        \caption{\samplername{}.}
        \label{fig:our_sampler}
    \end{subfigure}

    \begin{subfigure}[b]{\linewidth}
        \centering
        \includegraphics[
            width=\linewidth,
        ]{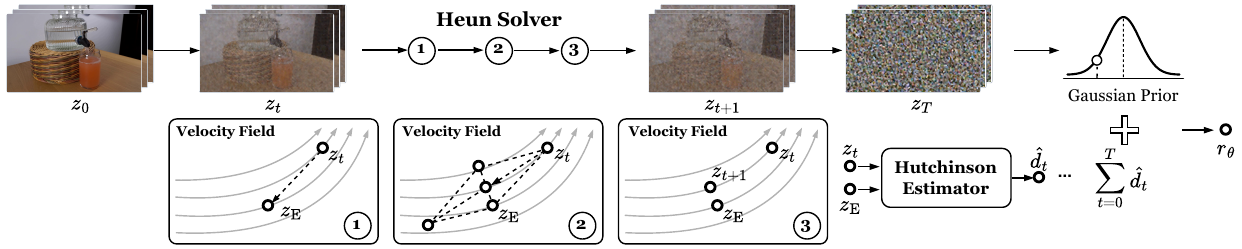}
        \caption{\rewardname{}.}
        \label{fig:our_reward}
    \end{subfigure}

    \caption{\textbf{Overview of our methods.}
    (a) \samplername{} uses an external loss on the predicted clean video to guide the perturbation noise. The guided noise is combined with the original endpoint prediction before denoising continues. Grey dashed lines denote the flow of gradients. 
    (b) \rewardname{} integrates the clean latent $\mathbf{z}_0$ to the noise endpoint $\mathbf{z}_1$ under the fixed base velocity field using a coarse Heun solver. Finite-difference Hutchinson estimator provides the divergence integral. In step 3, $z_t$ and $z_{t+1}$ fall on the same velocity line.
    }
    \label{fig:overview_of_our_methods}
\end{figure}

\subsection{Two-Dimensional Risk}
% \subsection{Hallucination Risk in Video Models}

%%
%% 幻觉风险应该分为两个层面，第一个层面是低概率密度幻觉风险，由于模型在训练时，拟合出来的样本分布是平滑的，因此会给多个模式之间的区域分配不可忽略的较低概率密度。
%% 所以在采样过程中会有一些case会落入这些低概率密度低区域，而这些区域在训练数据集中往往缺少support，因此可以认为具有很高的幻觉风险。

% 这一段简单概括一下
Directly quantifying hallucination in video generation is challenging, as it would require access to the unknown real-world conditional distribution $q^*$.
We therefore aim to formulate a unified theoretical framework for characterizing hallucination risk in video models. Inspired by the prior definition of the $\epsilon$-hallucination set, we define the $\delta$-low-density region as
\begin{equation}
\label{eq}
D_{\theta,\delta}(\mathbf{c}) = \{\mathbf{z_0} : p_\theta(\mathbf{z}_0 \mid \mathbf{c})\leq\delta\}.
\end{equation}
Here, we decompose the $\epsilon$-hallucination set into two parts:
\begin{equation}
    \label{eq:decompose_hallu_set}
    \begin{aligned}
        H_\epsilon(q^* \mid \mathbf{c})
        &=
        \Bigl(
            H_\epsilon(q^* \mid \mathbf{c})
            \cap
            D_{\theta,\delta}(\mathbf{c})
        \Bigr)
        \;\dot{\cup}\;
        \Bigl(
            H_\epsilon(q^* \mid \mathbf{c})
            \cap
            \overline{D_{\theta,\delta}(\mathbf{c})}
        \Bigr)
        \coloneqq
        H_{\mathrm{I}} \;\dot{\cup}\; H_{\mathrm{II}}.
    \end{aligned}
\end{equation}

This decomposition distinguishes two fundamentally different hallucination regimes according to whether the generated sample falls also into the low-density region in the model learned distribution.
$H_{\mathrm{I}}$ contains samples that have low-density in both $q^*$ and $p_\theta$.
% receive low probability density under both the real-world distribution $q^*$ and the learned distribution $p_\theta$.
In contrast, $H_{\mathrm{II}}$ contains samples that are hallucinate under $q^*$ but remain well supported by $p_\theta$.
Since \(q^*\) is inaccessible in practice, neither subset can be directly identified. However, low-density under $p_\theta$ is a necessary condition for membership in $H_\mathrm{I}$ and can be assessed using the model itself. 
This motivates us to characterize two layers of hallucination risk arising from sampling in low-density model regions and from drifting away from the real-world distribution despite high model density.
% This motivates us to characterize two distinct layers of hallucination risk, arising respectively from drifting away from the real-world distribution despite remaining in high-density regions of the model.

% 上面的这个decompose和下面的两层risk的衔接还需要优化一下，目前看起来联系不够紧密

\subsubsection{Low-Density Sampling Risk}

%% 第一层幻觉风险首先要明确是限制在模型学习到的分布上，使用不同的sampler对模型分布进行采样会得到不同的幻觉风险
%%
The first layer of hallucination risk is motivated by the decomposed set $H_{\mathrm I}$, which contains hallucinated samples that also fall into $D_{\theta,\delta}(\mathbf{c)}$.
Not every low-density sample is necessarily a hallucination. Nevertheless, mode interpolation provides an important mechanism through which generated samples may enter interpolated regions between learned modes~\citep{aithal2024understanding}. Therefore, We treat low model density as an model internal indicator of hallucination risk.
%is located in the model learned distribution.

% 下面这句话要修改一下
% Some regions in the learned distribution have low probability density, which may lead samplers to generate samples that fall into these regions.

\par\vspace{-3pt} % 缩小定义与上方正文的间距
\begin{restatable}[]{definition}{h}
    \label{def:low_density_risk}
    (\textbf{Low--Density Sampling Risk.})
    Let $p_\theta(\cdot \mid \mathbf{c})$ denote the learned model distribution and $p_{\theta,\mathcal{S}}(\cdot \mid \mathbf{c})$ the output distribution induced by a practical sampler $\mathcal{S}$. We define the sample-wise low-density sampling risk as
    \begin{equation}
        r_{\theta}(\mathbf{z}\mid \mathbf{c}) = -\log{p_\theta(\mathbf{z} \mid \mathbf{c})}.
    \end{equation}
    Then, the low-density sampling risk of Sampler $\mathcal{S}$ is defined as
    \begin{equation}
        \mathcal{R}_\theta(\mathcal{S}\mid \mathbf{c}) = \mathbb{E}_{\mathbf{z}\sim p_{\theta, \mathcal{S}}(\cdot \mid \mathbf{c})} [r_{\theta}(\mathbf{z}\mid \mathbf{c})] = \mathbb{E}_{\mathbf{z}\sim p_{\theta, \mathcal{S}}(\cdot \mid \mathbf{c})} [-\log{p_\theta(\mathbf{z}\mid \mathbf{c})}].
    \end{equation}
    % Equivalently, this risk is the cross-entropy $\mathcal{H}(p_{\theta,\mathcal{S}}, p_\theta)$, which can be decomposed as
    % \begin{equation}
    %     \mathcal{R}_\theta(\mathcal{S}\mid \mathbf{c}) = \mathcal{H}(p_{\theta,\mathcal{S}},p_\theta) = h(p_{\theta,\mathcal{S}}) + D_{\mathrm{KL}}(p_{\theta,\mathcal{S}}\Vert p_\theta).
    % \end{equation}

\end{restatable}
\par\vspace{-3pt} % 缩小定义与下方正文的间距

In our definition, the risk of a generated sample increases as the sample drops to the low-density region under $p_\theta$. The sampler-wise risk measures the overall tendency to generate outputs that are weakly supported by the learned model. 
% The composition further shows that the risk is not a pure divergence which may also decrease as sampling becomes concentrated in high-density regions.

% 这句话本来想解释这种的risk来源原因
% One of the main sources of hallucination risk come from mode interpolation.

% 具体的指标表征还需要再考虑一下

\subsubsection{High-Density Drifting Risk}
% \subsubsection{Model-Reality Mismatch Risk}

Although reducing low-density sampling risk favors samples with higher density under $p_\theta$, it only indicates that the model is more familiar with these samples. It does not guarantee that these samples are consistent with real-world dynamics.
The learned distribution $p_\theta$ may still differ from the real-world physical distribution $q^*$, creating a gap that we refer to as distribution drifting. We therefore introduce a second layer of risk to characterize this drift.

% problem: 目前这样定义看起来是不是有点简单了，可能会对第二层risk是否有必要建立在第一层的基础上（也就是说只衡量属于稳定模式区间的噪声的第二层risk）需要在解释上优化一下
% \par\vspace{-1pt} % 缩小定义与下方正文的间距
\begin{restatable}{definition}{mismatchrisk}
    \label{def:model_reality_mismatch_risk}
    \textbf{(High--Density Drifting Risk.)}
    Let $\ell_{\omega}(\mathbf{x},\mathbf{c}) \geq 0$ denote an external
    drifting loss, where a larger value indicates greater inconsistency
    between video $\mathbf{x}$ and real-world dynamics. For a sample with density above $\delta$ under $p_\theta$,
    $\mathbf{z}_0 \notin D_{\theta,\delta}(\mathbf{c})$, we define its
    sample-wise high-density drifting risk as
    \begin{equation}
        \label{eq:sample_mismatch_risk}
        r_{\omega}(\mathbf{z}_0 \mid \mathbf{c})
        =
        \ell_{\omega}
        \bigl(\mathcal{G}(\mathbf{z}_0),\mathbf{c}\bigr),
        \qquad
        \mathbf{z}_0 \notin D_{\theta,\delta}(\mathbf{c}).
    \end{equation}
    The high-density drifting risk of sampler $\mathcal{S}$ is defined as the expected external loss conditioned on the generated sample lying in a region where its density under $p_\theta$ exceeds $\delta$:
    \begin{equation}
        \label{eq:model_reality_mismatch_risk}
        \mathcal{R}_{\omega}(\mathcal{S} \mid \mathbf{c})
        =
        \mathbb{E}_{\mathbf{z}_0 \sim
        p_{\theta,\mathcal{S}}(\cdot \mid \mathbf{c})}
        \left[
            r_{\omega}(\mathbf{z}_0 \mid \mathbf{c})
            \,\middle|\,
            \mathbf{z}_0 \notin D_{\theta,\delta}(\mathbf{c})
        \right].
    \end{equation}
\end{restatable}
\par\vspace{-3pt} % 缩小定义与下方正文的间距

% This risk captures samples located in relatively high-density regions under $p_\theta$ but remain inconsistent with the real-world data manifold. Because $H_{\epsilon,\delta}^{\mathrm{II}} \subseteq \overline{D_{\theta,\delta}(\mathbf{c})}$, the conditional expectation measures mismatch only outside the
% low-density region. 
% Samples inside $D_{\theta,\delta}(\mathbf{c})$ are assessed by the low-density sampling risk, and excluding them here keeps the two sources of risk separate without assigning those samples zero mismatch.
% Since $q^*$ cannot be evaluated directly, $\ell_\omega$ provides an external estimate of the mismatch.

This risk measures the average external loss among samples with relatively high density under $p_\theta$. Conditioning on $\mathbf{z}_0 \notin D_{\theta,\delta}(\mathbf{c})$ follows the scope of $H_{\mathrm{II}}$ and measures the average severity of this distribution drifting among samples with relatively high model density. It does not imply that samples in $D_{\theta,\delta}(\mathbf{c})$ do not have external loss. Since $q^*$ is inaccessible, the accuracy of this assessment depends on how well $\ell_\omega$ captures this drift from real-world dynamics.

%% 上面这一部分把我们定义的理论交代清楚

%% 还需要设计一个toy example，体现layer-2 mismatch risk (已制作二维toy exp)

\subsection{From Two-Dimensional Risk to Practice}

\begin{wrapfigure}{r}{0.38\textwidth}
    % \vspace{-\intextsep}
    \centering
    \includegraphics[width=\linewidth]{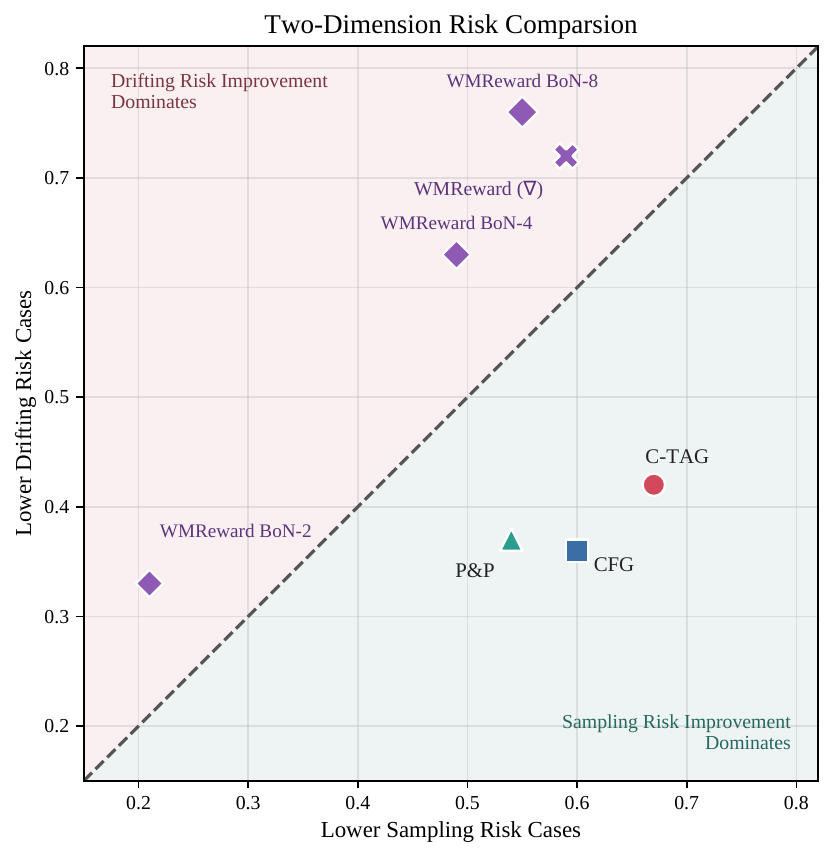}
    \caption{\textbf{Performance comparison across the two-dimensional risk.} Each axis reports the fraction of 100 cases with lower corresponding risk than the Euler sampler.}
    \label{fig:risk_comparison}
\end{wrapfigure}

% Figure~\ref{fig:risk_comparison} shows that different methods
% exhibit distinct improvement patterns across the two risk dimensions.

% \subsection{Hallucination-Aware Video Sampling}
%% 这部分
%% 首先是讲述为什么pnp和tag是能够降低layer-1 hallucination risk的
%% 以及，为什么只有reward guidance是不足够的，也就是说，当采样接近高密度区域时，reward guidance才能取得更大收益，也就是layer-2 hallucination risk是有意义的，但是必须建立在layer-1的基础上
We use a first-order Euler solver as a reference sampler, denoted by $\mathcal{S}_{\mathrm E}$. All sampler comparisons are conducted under the same condition $\mathbf{c}$. Previous work has shown that latent refinement exhibits mode-seeking behavior~\citep{jang2026selfrefining}, while gradient guidance uses an external reward to steer generation toward more physically plausible outcomes~\citep{yuan2026wmreward}. 
Mathematically, we interpret latent refinement through its idealized low-temperature Langevin counterpart at a fixed noise level, while gradient guidance approximates a reward-reweighted target distribution. 
We provide a detailed mathematical analysis in Appendix~\ref{app:risk_mechanisms}.
~\Cref{fig:risk_comparison} shows that latent refinement used in P\&P reduces low-density sampling risk in more cases than high-density drifting risk, while WMReward shows the reverse pattern.
Under our framework, latent refinement addresses low-density sampling risk by favoring regions of higher model density, while gradient guidance provides a way to address high-density drifting through the external guidance. 
These complementary roles motivate the incorporation of gradient guidance into the refinement process to address both risks within each iteration.

%% 关键在于讲述，当减少low-density risk时，reward guidance才能取得更大收益 
%% 这样才能说明为什么mismatch risk为什么一定要建立在low-density的基础上
%% 真实流形分布与稳定mode距离较近(?)

%% 目前有一个角度是，说明low-density risk与predict endpoint的联系
%% 目前已有的实验结论是： 对于pnp以及tag这样的方法，在做完latent优化后，立即计算noisy-marginal CNF，均出现下降，说明p_t(z_t|c) 上升，并且

\subsubsection{Hallucination-Aware Video Sampling}
% \subsubsection{Hallucination-Aware Video Sampling Algorithm}

%% 依照我们的理论引入两种改进算法
%% 改进算法的创新机制放进appendix具体讲解

\begin{figure}[t]
    \centering
    \begin{subfigure}[b]{0.49\linewidth}
        \centering
        \includegraphics[
            width=\linewidth,
            trim={0mm 0mm 7mm 4mm},
            clip
        ]{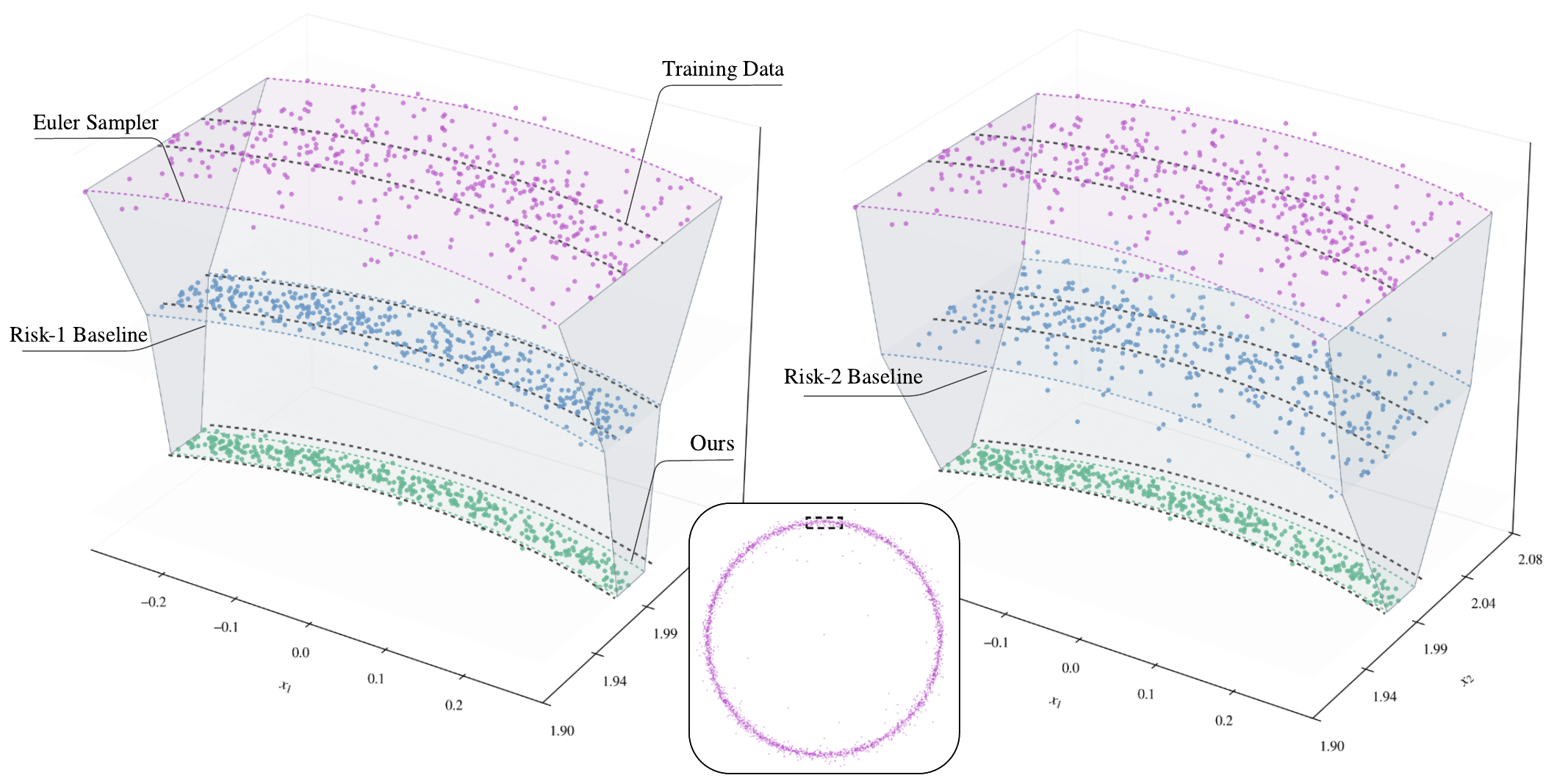}
        \caption{Circle distribution.}
        \label{fig:first}
    \end{subfigure}
    \hfill
    \begin{subfigure}[b]{0.49\linewidth}
        \centering
        \includegraphics[
            width=\linewidth,
            trim={0mm 0mm 7mm 4mm},
            clip
        ]{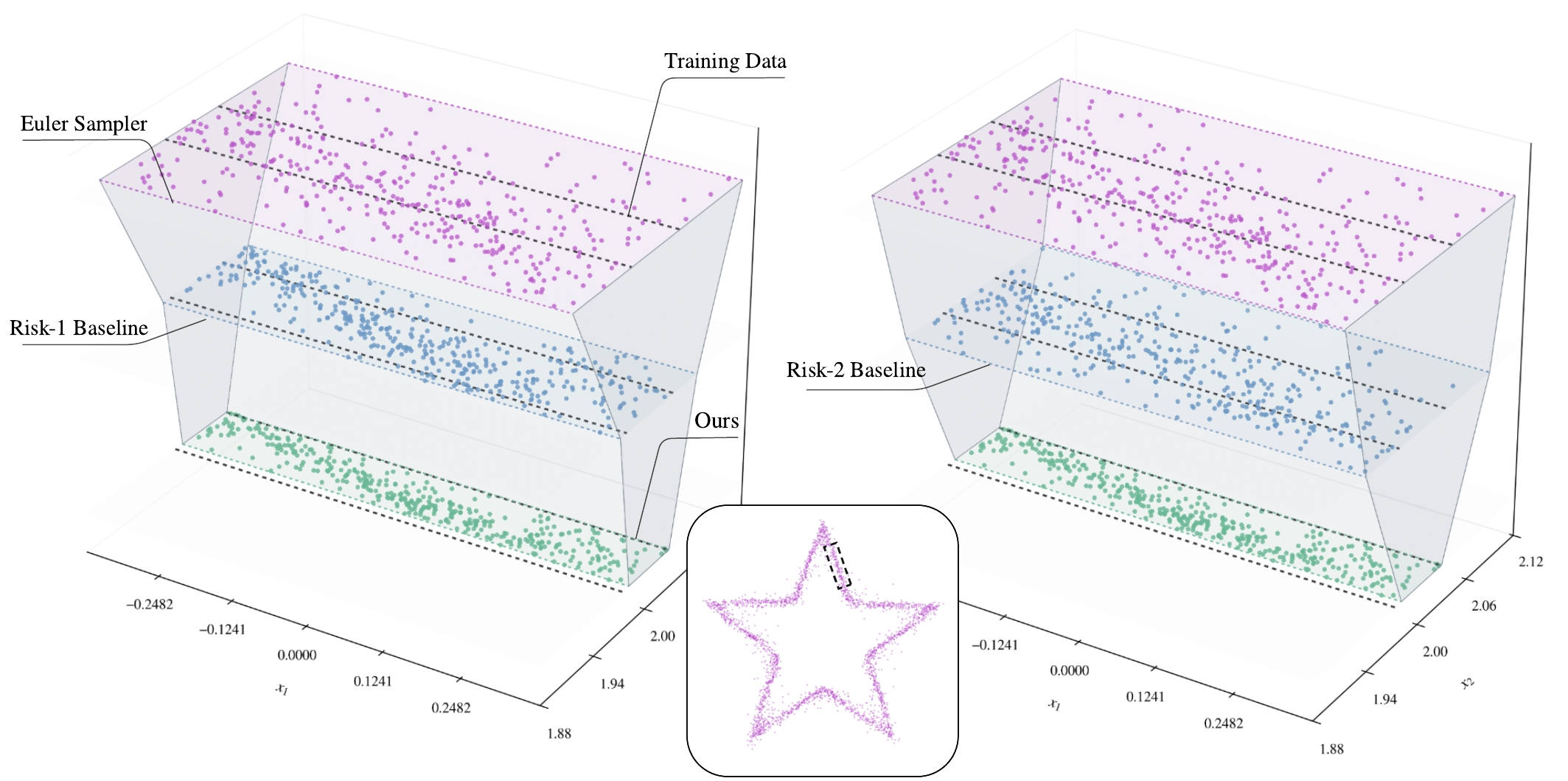}
        \caption{Star distribution.}
        \label{fig:second}
    \end{subfigure}

    \caption{\textbf{Sampling results on 2D toy distributions.}
    The stacked layers compare Euler sampler, P\&P (risk-1 baseline),  guidance based on the negative log-likelihood (NLL) under the true data distribution (risk-2 baseline), and~\samplername. 
    The Euler sampler uses a large number of sampling steps to reduce discretization error.
    Purple reference curves indicate the training-data geometry.}
    \label{fig:comparison}
\end{figure}

Building on our two-layer hallucination risk framework, we introduce~\samplername{}, a training-free sampler designed to address low-density sampling risk and high-density drifting risk within the same refinement process. At each iteration, gradient guidance adjusts the perturbation noise, and the generative model re-predicts the clean endpoint from the guided perturbation. The sampling procedure is illustrated in ~\Cref{fig:overview_of_our_methods}(a). We further illustrate this design through two-dimensional toy experiments with circle- and star-shaped data distributions in ~\Cref{fig:comparison}. We compare~\samplername{} with Euler sampling and baselines that target either low-density sampling risk or high-density drifting risk alone. In the displayed regions, our samples are more tightly concentrated around the training-data reference curves than those from Euler or either baseline.

\begin{table*}[t]
\centering
\begin{adjustbox}{width=\linewidth}
\begin{tabular}{lccccc}
\toprule
\multirow{3}{*}{\textbf{Method}}
& \multicolumn{4}{c}{\textbf{Physics-IQ Metrics}} 
& \multirow{3}{*}{\textbf{Score} $\uparrow$} \\
\cmidrule(lr){2-5}
& \makecell{\textbf{Spatial}\\\textbf{IoU} $\uparrow$}
& \makecell{\textbf{Spatiotemporal}\\\textbf{IoU} $\uparrow$}
& \makecell{\textbf{Weighted}\\\textbf{Spatial IoU} $\uparrow$}
& \makecell{\textbf{MSE}\\\textbf{Score} $\uparrow$}
& \\
% \midrule

% \multicolumn{6}{@{}c@{}}{\textit{Reported reference results}} \\
% Pika~\TODO{add citation}
% & -- & -- & -- & -- & 13.00$^{\dagger}$ \\
% Runway Gen-3
% & -- & -- & -- & -- & 22.80$^{\dagger}$ \\
% Sora 2
% & -- & -- & -- & -- & 26.50$^{\dagger}$ \\
% MAGI-1
% & -- & -- & -- & -- & 30.20$^{\dagger}$ \\

\midrule
SANA-Video2.0~\citep{chen2026sanavideo2}
& 0.2209 & 0.1986 & 0.1309 & 0.0630 & 15.33 \\
CogVideoX-5B~\citep{yang2025cogvideox}
& 0.3686 & 0.3806 & 0.2286
& 0.2971 & 31.87 \\
Wan2.2-TI2V-5B~\citep{wan2025}
& 0.3295 & 0.2442 & 0.1727
& 0.3074 & 26.34 \\
SkyReels-V2-I2V~\citep{chen2025skyreelsv2}
& 0.2717 & 0.2642 & 0.1917
& 0.1084 & 20.90 \\

\midrule
SkyReels-V2-I2V
& 0.2717 & 0.2642 & 0.1917 & 0.1084 & 20.90 \\
\quad $+$ P\&P~\citep{jang2026selfrefining}
& \underline{0.2974} & \underline{0.2685} & \underline{0.2157}
& \textbf{0.1365} & \underline{22.95} \\
\quad $+$ WMReward ($\nabla$)~\citep{yuan2026wmreward}
& 0.2742 & 0.2640 & 0.1939 & 0.1100 & 21.05 \\
\rowcolor{gray!10}
\quad $+$ \textbf{\samplername}~(ours)
& \textbf{0.3132} & \textbf{0.2750} & \textbf{0.2342}
& \underline{0.1289} & \textbf{23.78} \\

\midrule
Wan2.2-TI2V-5B
& 0.3295 & \textbf{0.2442} & 0.1727
& \textbf{0.3074} & 26.34 \\
\quad $+$ P\&P~\citep{jang2026selfrefining}
& \underline{0.3796} & 0.2322 & \underline{0.2109}
& 0.2916 & \underline{27.86} \\
\quad $+$ WMReward ($\nabla$)~\citep{yuan2026wmreward}
& 0.3310 & \underline{0.2427} & 0.1732
& \underline{0.3072} & 26.35 \\
\rowcolor{gray!10}
\quad $+$ \textbf{\samplername}~(ours)
& \textbf{0.3927} & 0.2360 & \textbf{0.2243}
& 0.2846 & \textbf{28.44} \\

\bottomrule
\end{tabular}
\end{adjustbox}

\caption{\textbf{Comparison of different samplers on Physics-IQ.}
We report the Physics-IQ Verified metrics and overall score.
% General video models are included as references.
The lower blocks compare inference-time methods using Wan2.2-TI2V-5B
and SkyReels-V2-I2V. 
% No candidate selection is used.
% Method references:
% Wan2.2~\citep{wan2025},
% SANA-Video2.0~\TODO{cite},
% SkyReels-V2~\citep{chen2025skyreelsv2},
% P\&P~\citep{jang2026selfrefining},
% and WMReward~\citep{yuan2026wmreward}.
$\nabla$ denotes gradient guidance of WMReward.}
% $^{\dagger}$ denotes results reported in prior work rather than evaluated by us.}
\label{tab:physics_iq_nonbon}
\end{table*}

\paragraph{Sampling procedure}
At noise level $t$, we perform $K_t$ refinement iterations, starting from the current predicted clean endpoint $\hat{\mathbf{z}}_{0\mid t}$.
Each iteration samples perturbation noise $\boldsymbol{\epsilon}$ to construct $\mathbf{u}_t$ and predict its clean endpoint $\hat{\mathbf{u}}_{0\mid t}$.
We compute the external loss $\mathcal{L} = \ell_\omega(\mathcal{G}(\hat{\mathbf{u}}_{0\mid t}), \mathbf{c})$ and its gradient with respect to $\boldsymbol{\epsilon}$.
After root mean square (RMS) normalization, we update the noise with step size $\eta_t$.
We then construct $\mathbf{u}_t^{+}$ with the updated noise and re-evaluate the model to obtain $\hat{\mathbf{z}}_{0\mid t}^{+}$, which becomes the input to the next iteration.
% A superscript $+$ denotes quantities after the guidance update.
Algorithm~\ref{alg:guided_latent_refine} in Appendix~\ref{app:algorithms} presents one refinement iteration.

\subsubsection{Hallucination-Aware Scoring}
% \subsubsection{Estimating Low-density Sampling Risk}
%% 用于引入代理指标，计算第一层risk
%% 创新点1: 我们使用CNF指标来评估不同的sampler对layer-1 hallucination risk的降低程度
%% 有可能可以使用计算cost更低的DAE来代替

To estimate hallucination risk $r_\theta$ in Definition~\ref{def:low_density_risk}, we integrate the fixed base ODE $\frac{\mathrm{d}\mathbf{z}_t}{\mathrm{d}t} = v_\theta(\mathbf{z_t}, t, \mathbf{c})$ from the clean latent $\mathbf{z}_0$ to its noise endpoint $\mathbf{z}_1$. The continuous normalizing flow (CNF) change-of-variables formula~\citep{chen2018neural} gives the following
\begin{equation}
\label{eq:cnf}
    -\log{p_\theta(\mathbf{z}_0\mid\mathbf{c})} = -\log{\varphi(\mathbf{z}_1)} - \int_0^1\nabla_{\mathbf{z}_t}\cdot v_\theta(\mathbf{z}_t, t, \mathbf{c}) dt,
\end{equation}
where $\varphi$ is the standard Gaussian density.
The first term is the negative log-likelihood (NLL) under the Gaussian prior, whereas the integral accounts for volume changes along the flow.
Computing the divergence is expensive for high-dimensional video latent.
We estimate the divergence using Hutchinson's estimator~\citep{hutchinson1989stochastic}, with forward finite differences to avoid backpropagation through the video model:
\begin{equation}
\label{eq:hutchinson_estimator}
\hat{d}_t = \frac{1}{P} \sum_{p=1}^{P} \mathbf{r}_p^\top \mathbf{J}_{\theta,t} \mathbf{r}_p \approx \frac{1}{P \delta_t} \sum_{p=1}^{P} \mathbf{r}_p^\top \bigl[v_\theta(\mathbf{z}_t + \delta_t \mathbf{r}_p, t, \mathbf{c}) - v_\theta(\mathbf{z}_t, t, \mathbf{c})\bigr].
\end{equation}
Here, $\mathbf{J}_{\theta,t}$ is the Jacobian of $v_\theta$ with respect to $\mathbf{z}_t$.
The random probes $\mathbf{r}_p$ are sampled independently and reused across integration steps.
The perturbation scale $\delta_t$ is computed from $\delta_{\mathrm{rel}}$ and the latent root mean square.
We use Heun's second-order method~\citep{karras2022elucidating} on a coarse time grid to compute the inversion trajectory and accumulate the finite-difference divergence estimates.
We refer to this evaluation as \rewardname{}.
An overview of \rewardname{} is shown in ~\Cref{fig:overview_of_our_methods} (b), with the complete procedure given in ~\Cref{alg:sparse_cnf} (Appendix~\ref{app:algorithms}).

\begin{table*}[t]
\centering
\begin{adjustbox}{width=\linewidth}
\begin{tabular}{lccccc}
\toprule
\multirow{3}{*}{\textbf{Method}}
& \multicolumn{4}{c}{\textbf{Physics-IQ Metrics}}
& \multirow{3}{*}{\textbf{Score} $\uparrow$} \\
\cmidrule(lr){2-5}
&
\makecell{\textbf{Spatial}\\\textbf{IoU} $\uparrow$}
& \makecell{\textbf{Spatiotemporal}\\\textbf{IoU} $\uparrow$}
& \makecell{\textbf{Weighted}\\\textbf{Spatial IoU} $\uparrow$}
& \makecell{\textbf{MSE}\\\textbf{Score} $\uparrow$}
& \\
\midrule

\multicolumn{6}{@{}c@{}}{\textit{Base sampler}} \\
Random
& 0.3372 & 0.1828 & 0.2418 & 0.2936 & 26.38 \\

Qwen3-VL~\citep{bai2025qwen3vl}
& \underline{0.3449} & \textbf{0.2572} & 0.2161 & 0.3091 & 28.18 \\

Gemma-4~\citep{gemmateam2026gemma4}
& 0.3380 & \underline{0.2392} & 0.1847 & 0.3002 & 26.55 \\

WMReward~\citep{yuan2026wmreward}
& 0.3175 & 0.1972 & \textbf{0.3647} & \textbf{0.3410} & \underline{30.51} \\

GeoPhys~\citep{interno2026geophys}
& 0.3278 & 0.1782 & 0.2307 & 0.2900 & 25.67 \\

% Sparse-CNF (ours)
% & 0.3492 & 0.2074 & 0.2933 & 0.3227 & 29.31 \\
% Hybrid
% & 0.3515 & 0.2172 & 0.3312 & 0.3353 & 30.88 \\

\rowcolor{gray!10}
\textbf{\rewardname}~(ours)
& \textbf{0.3515} & 0.2172 & \underline{0.3312}
& \underline{0.3353} & \textbf{30.88} \\

\midrule
\multicolumn{6}{@{}c@{}}{\textit{P\&P sampler}} \\
Random
& 0.3647 & 0.2308 & 0.2001 & 0.2852 & 27.02 \\

Qwen3-VL~\citep{bai2025qwen3vl}
& \underline{0.3910} & 0.2368 & 0.2197 & 0.3020 & 28.74 \\

Gemma-4~\citep{gemmateam2026gemma4}
& 0.3455 & 0.2257 & 0.1897 & 0.2888 & 26.24 \\

WMReward~\citep{yuan2026wmreward}
& 0.3681 & \textbf{0.3025} & \underline{0.2304}
& \underline{0.3246} & \underline{30.64} \\

GeoPhys~\citep{interno2026geophys}
& 0.3757 & 0.2049 & 0.2058 & 0.2925 & 26.97 \\

% Sparse-CNF (ours)
% & 0.4002 & 0.2739 & 0.2383 & 0.3436 & 31.40 \\
% Hybrid
% & 0.3917 & 0.2384 & 0.2910 & 0.3349 & 31.40 \\

\rowcolor{gray!10}
\textbf{\rewardname}~(ours)
& \textbf{0.3917} & \underline{0.2384} & \textbf{0.2910}
& \textbf{0.3349} & \textbf{31.40} \\

\midrule
\multicolumn{6}{@{}c@{}}{\textit{\samplername}} \\
Random
& 0.3681 & 0.2020 & 0.2325 & 0.2846 & 27.18 \\

Qwen3-VL~\citep{bai2025qwen3vl}
& \underline{0.3777} & \underline{0.2385} & 0.2178 & 0.3057 & 28.49 \\

Gemma-4~\citep{gemmateam2026gemma4}
& 0.3652 & 0.2354 & 0.2095 & 0.2958 & 27.65 \\

WMReward~\citep{yuan2026wmreward}
& 0.3761 & 0.2303 & \textbf{0.3294}
& \underline{0.3315} & \underline{31.68} \\

GeoPhys~\citep{interno2026geophys}
& 0.3686 & 0.2011 & 0.2190 & 0.2907 & 26.98 \\

% Sparse-CNF (ours)
% & 0.3828 & 0.2257 & 0.2798 & 0.3385 & 30.67 \\
% Hybrid
% & 0.3965 & 0.2443 & 0.3127 & 0.3364 & 32.25 \\

\rowcolor{gray!10}
\textbf{\rewardname}~(ours)
& \textbf{0.3965} & \textbf{0.2443} & \underline{0.3127}
& \textbf{0.3364} & \textbf{32.25} \\

\bottomrule
\end{tabular}
\end{adjustbox}

\caption{\textbf{Results of scorer methods on Physics-IQ with Best-of-$N$ generation.}
Random denotes selecting a candidate uniformly from the $N=8$ candidates pool.
Candidate generation uses Wan2.2-TI2V-5B.
% The results show that our propose~\rewardname~achieves the best score
% compared with SOTA baselines, indicating the effectiveness of our theory
% on selecting the good generated samples.\TODO{refine}
% \rewardname~selects the candidate with the lowest estimated negative
% log-density and the results show that xxx.
}
\label{tab:physics_iq_bon}
\end{table*}

\section{Experiments}

\subsection{Experimental Setup}

\label{sec:exp_setup}

\paragraph{Implementation details}
We conducted all experiments on flow-matching-based video generative models,
including Wan2.2-TI2V-5B and SkyReels-V2-I2V-1.3B.
For VLM-based scoring, we use Qwen3-VL-8B-Instruct~\citep{bai2025qwen3vl} and Gemma-4-E4B-it~\citep{gemmateam2026gemma4}. 
Details of the scoring procedure and results with other VLMs are provided in Appendix~\ref{app:vlm_results}.
In practice, we use the WMReward surprise score~\citep{yuan2026wmreward} as the external loss.
For \rewardname{}, the default settings are $H=8$ Heun steps, $P=2$ Hutchinson probes, and a relative perturbation scale of $\delta_{\mathrm{rel}}=0.0125$.
% Following \citet{jang2025dreamgen}, we ask the model to assess physical plausibility for the input description and specify a binary 0/1 response.
% Following the answer-likelihood approach of \citet{lin2024evaluating}, we use the log-probability of the positive answer token 1 as the reward.
% For a fair comparison, we use the same sampling budget for all Best-of-N (BoN) selection methods and set \(N=8\). 
% That is, each BoN method generates eight candidate videos for each input condition and selects one video according to its reward.
We compare scoring methods using a best-of-$N$ (BoN) protocol, where each method selects its top-ranked candidate from $N = 8$ video candidates generated per input condition.
Comparisons with $N \in \{2, 4, 8\}$ are provided in Appendix~\ref{app:bon-details} to examine how performance changes with the number of candidates.
All experiments are conducted on the RTXA6000 platform.

% \TODO{zheng, 9.23}

\paragraph{Baselines and benchmarks}
% We compared our approach against four strong baselines, including the base model without any extra strategy, P\&P~\cite{jang2026selfrefining}, WMReward~\cite{yuan2026wmreward} and Phys4D~\cite{lu2026phys4d}. Detailed description of these methods can be found in Appendix~\ref{app:baselines}.
For sampling, we compare with the base model, P\&P~\citep{jang2026selfrefining}, and WMReward ($\nabla$)~\citep{yuan2026wmreward}. We also include other physics-aware methods in the T2V evaluation. 
For candidate selection, we compare against WMReward~\citep{yuan2026wmreward}, GeoPhys~\citep{interno2026geophys}.
We choose these baselines for their competitive and SOTA performance.
% and because each primarily addresses one of the two risks defined in our framework.
We evaluate I2V generation on Physics-IQ~\citep{radsch2026physicsiqverified} using the Verified metrics and T2V generation on VideoPhy~\citep{bansal2025videophy}.
Details are provided in Appendix~\ref{app:baselines-benchmarks}.

% \TODO{refine, describe the reasons why select these methods}

% \TODO{wu, 9.23}

% \paragraph{Benchmarks} 
% % We evaluate physics alignment using the Physics-IQ benchmark~\cite{motamed2026physicsiq},
% % a real-world video benchmark designed to measure whether generative video models can predict physically plausible future dynamics.
% % This benchmark contains 396 high-quality videos, each 8 seconds long, covering 66 scenarios across 5 categories.
% % Each scenario is recorded from three static camera viewpoints and repeated twice under the same conditions.
% We evaluate physical plausibility on Physics-IQ~\cite{motamed2026physicsiq} for image-to-video (I2V) generation and VideoPhy~\TODO{cite} for text-to-video (T2V) generation. For Physics-IQ, we report results using the Verified metrics. Physics-IQ is a benchmark that contains 396 high-quality videos, covering 66 scenarios across 5 categories. For VideoPhy, we use the test split of 344 prompts covering physical interactions and report semantic adherence (SA) and physical commonsense (PC).
% Benchmark details and evaluation protocols are provided in Appendix\TODO{ref}.

% \TODO{refine}

% \TODO{wu, 9.23}

\begin{figure}[t]
    \centering
    \begin{minipage}[t]{0.49\linewidth}
        \vspace{0pt}
        \centering
        {\footnotesize\itshape
        ``...brown tennis ball rolls from the left straight out of the black tube and hits the olive green fabric box...''\par}
        \vspace{2pt}
        \includegraphics[width=\linewidth]
        {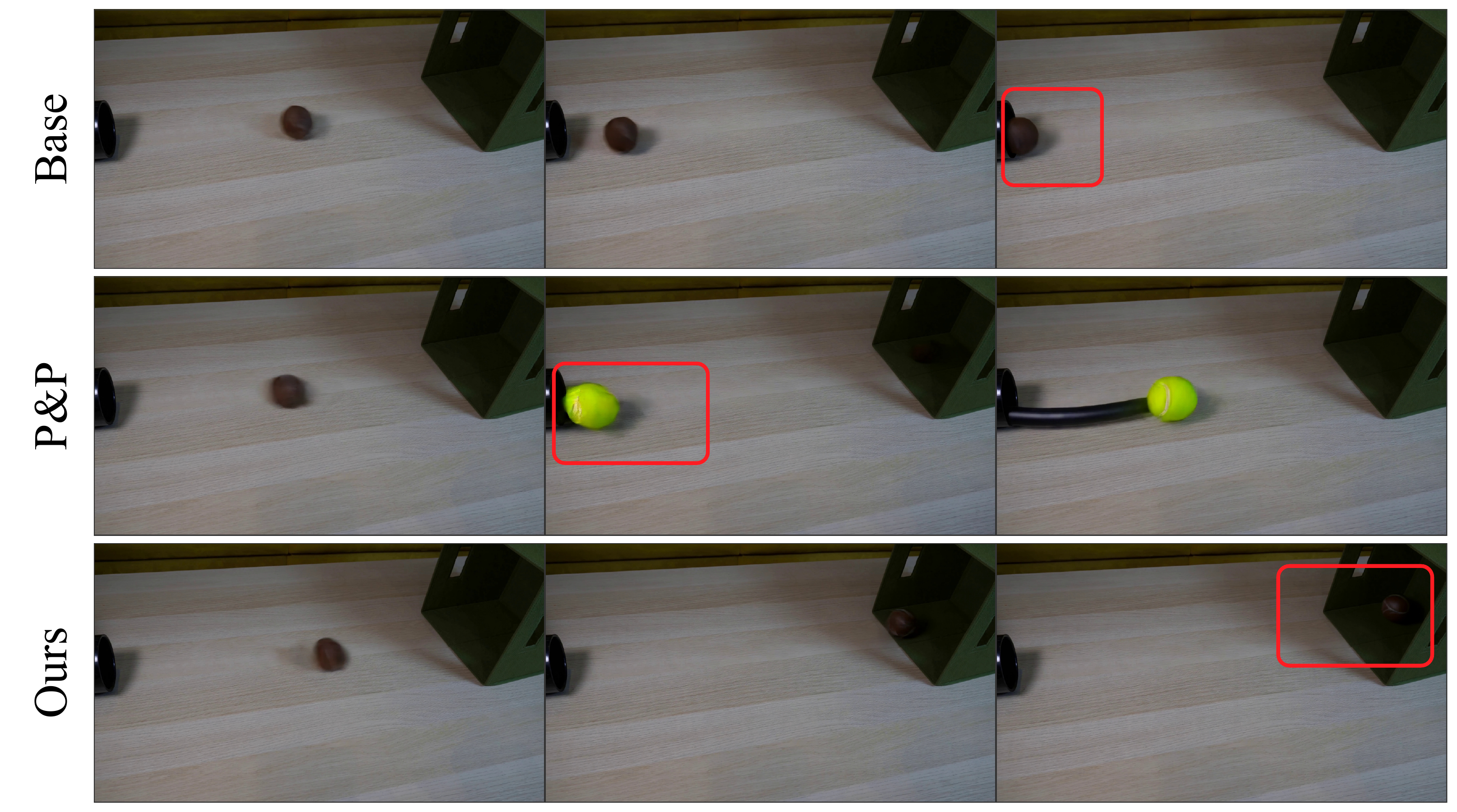}
    \end{minipage}\hfill
    \begin{minipage}[t]{0.49\linewidth}
        \vspace{0pt}
        \centering
        {\footnotesize\itshape
        ``...yellow rubber duck sits stationary on a light brown wooden table...''\par}
        \vspace{2pt}
        \includegraphics[width=\linewidth]
        {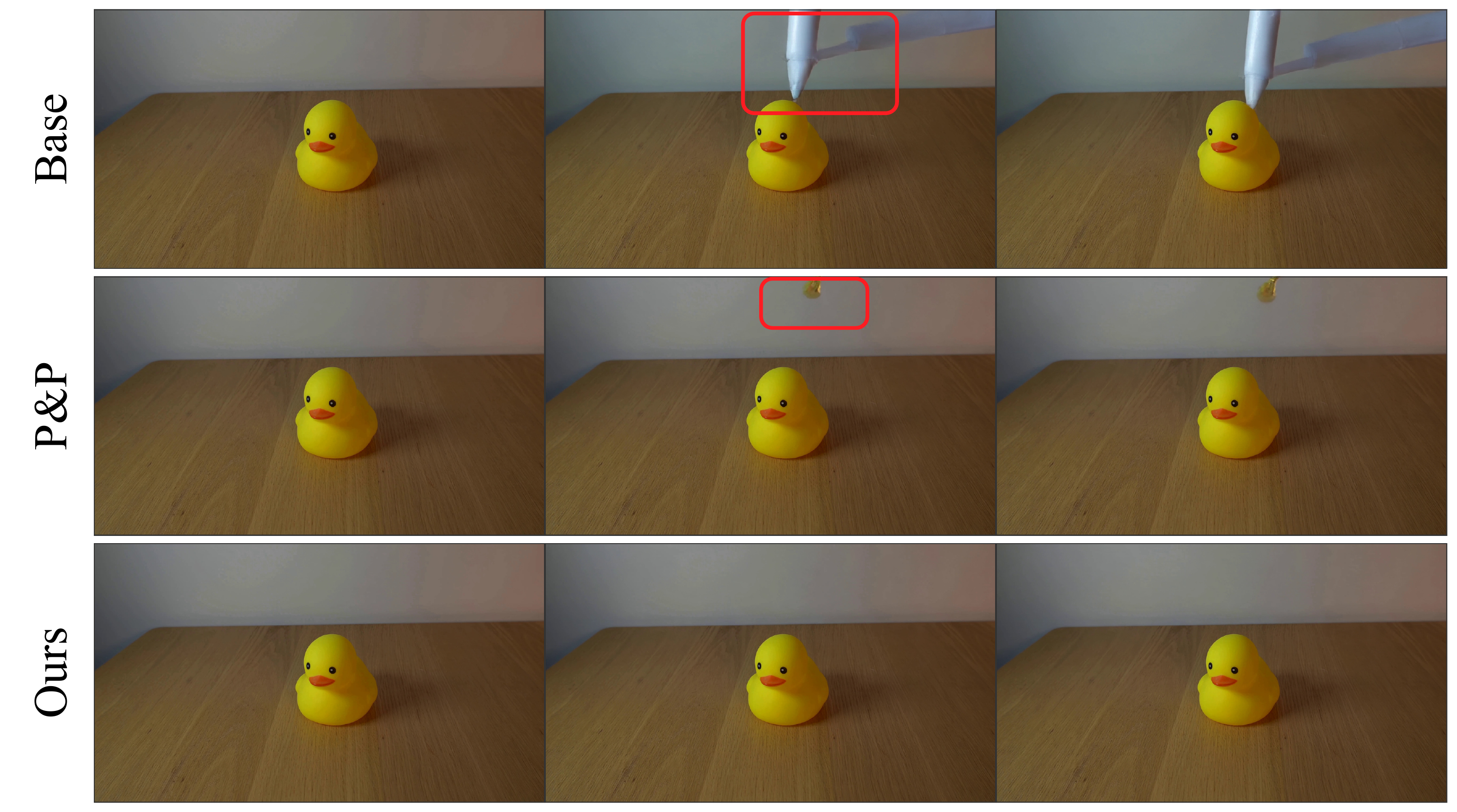}
    \end{minipage}

    \caption{\textbf{Qualitative comparison of samplers on Physics-IQ.}
   We compare sampling methods in a dynamic collision scene (left) and a static scene (right).
\samplername{} produces more physically plausible results than the base sampler and P\&P in both cases.}
    \label{fig:comp_sampling}
\end{figure}

\begin{table*}[t]
\centering
\small
% \setlength{\tabcolsep}{3pt}
% \renewcommand{\arraystretch}{1.10}

% \begin{tabularx}{\linewidth}{@{}>{\raggedright\arraybackslash}X *{4}{c}@{}}
\begin{adjustbox}{width=0.92\linewidth}
\begin{tabular}{lcccc}

\toprule
\multirow{1}{*}{\textbf{Method}}
% & \multicolumn{4}{c}{\textbf{Physical Commonsense (PC)} $\uparrow$} \\
% \cmidrule(lr){2-5}
% & \makecell{\textbf{Solid--}\\\textbf{Solid}}
% & \makecell{\textbf{Solid--}\\\textbf{Fluid}}
% & \makecell{\textbf{Fluid--}\\\textbf{Fluid}}
% & \textbf{Overall} \\
& \textbf{Solid--Solid} $\uparrow$
& \textbf{Solid--Fluid} $\uparrow$
& \textbf{Fluid--Fluid} $\uparrow$
& \textbf{Overall} $\uparrow$ \\
\midrule

\multicolumn{5}{@{}l@{}}{\textit{General video models}} \\
\addlinespace[3pt]
Wan2.1-T2V-1.3B~\citep{wan2025}
& 0.18 & 0.24 & 0.27 & 0.22 \\
SanaVideo2.0~\citep{chen2026sanavideo2}
& \textbf{0.27} & 0.23 & \underline{0.33} & \underline{0.26} \\
Wan2.2-TI2V-5B~\citep{wan2025}
& 0.21 & \textbf{0.32} & 0.25 & \underline{0.26} \\

\midrule
\multicolumn{5}{@{}l@{}}{\textit{Physics-aware methods}} \\
\addlinespace[3pt]
DreamWorld~\citep{tan2026dreamworld}
& 0.16 & 0.29 & 0.31 & 0.24 \\
VPT~\citep{zheng2026enhancing}
& 0.17 & 0.24 & \underline{0.33} & 0.22 \\
P\&P~\citep{jang2026selfrefining}
& \underline{0.22} & 0.23 & \underline{0.33} & 0.24 \\
WMReward ($\nabla$)~\citep{yuan2026wmreward}
& 0.21 & \textbf{0.32} & 0.24 & \underline{0.26} \\

\addlinespace[3pt]
\rowcolor{gray!10}
\textbf{\samplername}~(ours)
& \textbf{0.27} & \underline{0.30} & \textbf{0.35} & \textbf{0.30} \\

\bottomrule
\end{tabular}
\end{adjustbox}

\caption{\textbf{Comparison of different samplers on VideoPhy.}
We report category-wise and overall Physical Commonsense (PC).
Higher scores are better. Our method uses Wan2.2-TI2V-5B as the base model.
$\nabla$ denotes gradient guidance of WMReward.}
\label{tab:videophy_benchmark}
\end{table*}

\subsection{Comparison with Existing Methods}

\subsubsection{Quantitative Results}

\noindent\textbf{Sampler comparison.}
Table~\ref{tab:physics_iq_nonbon} compares inference-time sampling methods on Physics-IQ Benchmark. On Wan2.2-TI2V-5B, ~\samplername achieves the highest overall score among compared samplers, exceeding the baseline by \textbf{+2.10}, and leads in Spatial IoU and Weighted Spatial IoU.
On SkyReels-V2-I2V, our method also achieves the highest overall score, exceeding the baseline by \textbf{+2.88}.
These results suggest that latent refinement and gradient guidance complement each other, producing consistent improvements over either method alone across both backbones.
Table~\ref{tab:videophy_benchmark} reports results on VideoPhy.
\samplername{} achieves the highest overall PC score among the compared physics-aware methods. We further compare the two hallucination risks across samplers in Appendix~\ref{app:risk_comparison}.

\noindent\textbf{Scorer comparison.}
Table~\ref{tab:physics_iq_bon} compares scorer methods for best-of-$N$ selection with $N = 8$ using Wan2.2-TI2V-5B. \rewardname{} achieves the highest overall score under all three samplers.
Compared with random selection, \rewardname{} improves the overall score by \textbf{+4.50}, \textbf{+4.38}, and \textbf{+5.07} under the base, P\&P, and \samplername{}, respectively. Our scorer also leads in Spatial IoU across all three samplers and ranks first in three of the four metrics under both P\&P and \samplername{}. These gains show that our scorer effectively selects more physically plausible videos and remains beneficial even when candidate generation is improved through \samplername.

% \TODO{zheng, 9.23}

\subsubsection{Qualitative Results}

\noindent\textbf{Sampler comparison.}
~\Cref{fig:comp_sampling} compares sampling methods in dynamic and static scenes.
In the left case, the base sampler moves the ball back toward the tube, while P\&P changes the ball's color from brown to yellow and introduces an additional tube. \samplername{} preserves the appearance of the original ball and generates a trajectory into the box. In the right case, both baselines introduce extraneous objects above the duck, whereas \samplername{} preserves the stationary scene.

\begin{figure}[t]
    \centering
    \begin{minipage}[t]{0.49\linewidth}
        \vspace{0pt}
        \centering
        {\footnotesize\itshape
        ``...brown tennis ball rolls straight out of the black pipe
        and hits the rubber duck...''\par}
        \vspace{2pt}
        \includegraphics[width=\linewidth]
        {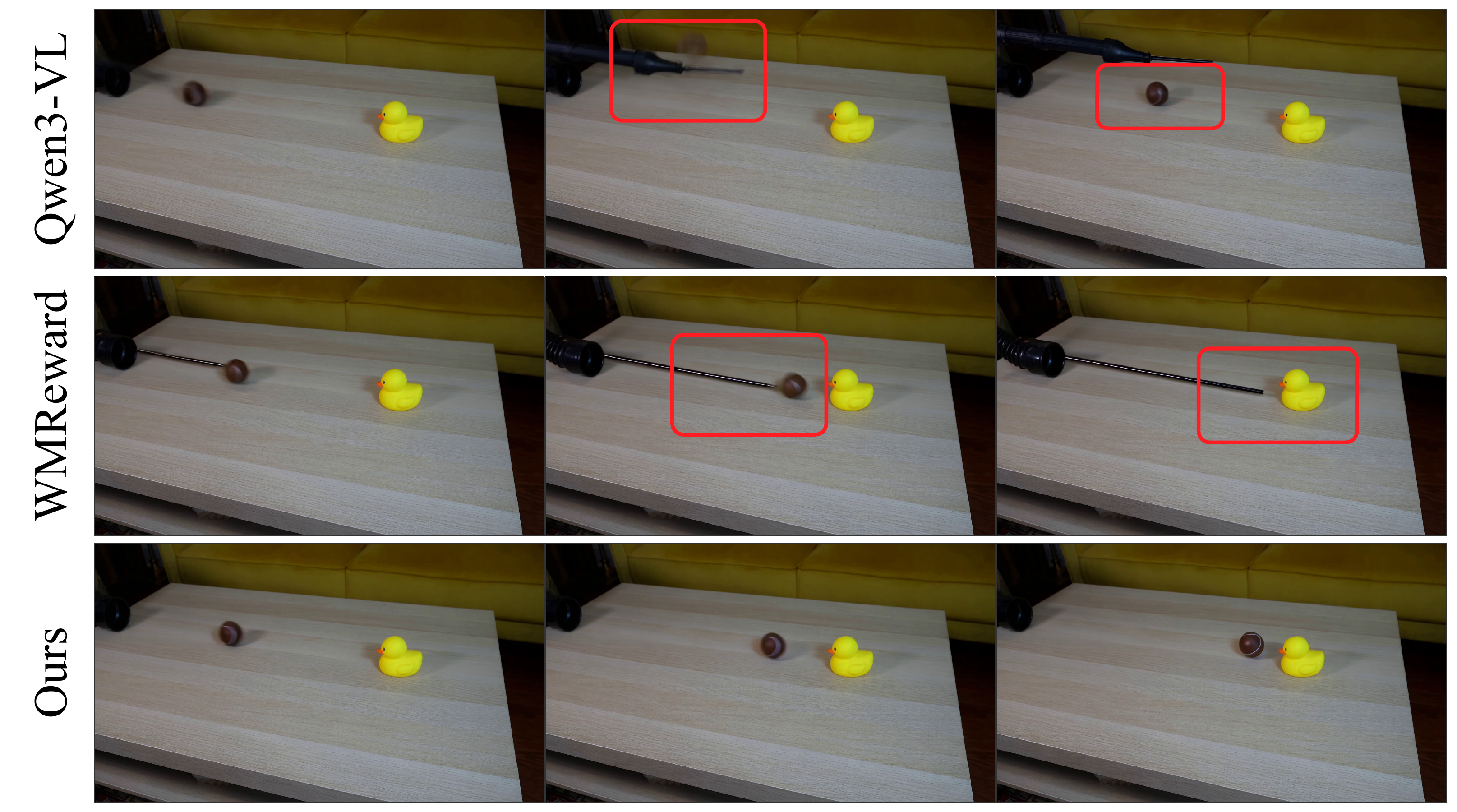}
    \end{minipage}\hfill
    \begin{minipage}[t]{0.49\linewidth}
        \vspace{0pt}
        \centering
        {\footnotesize\itshape
        ``...blue-black grabber tool releases the blue tennis ball
        above a pile of green kinetic sand...''\par}
        \vspace{2pt}
        \includegraphics[width=\linewidth]
        {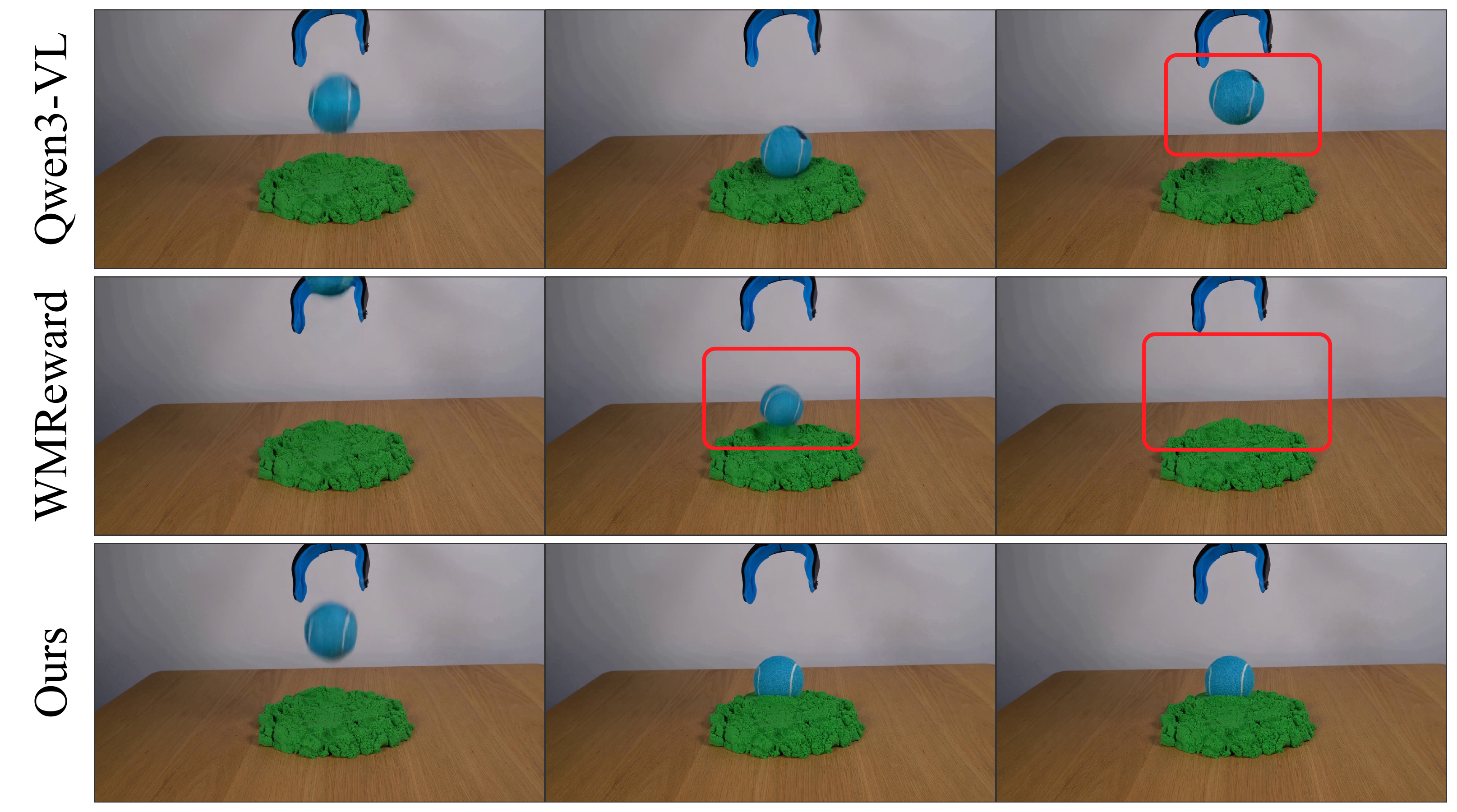}
    \end{minipage}

    \caption{\textbf{Qualitative comparison of candidates selection on Physics-IQ.}
    For each prompt, all scorer methods select from the same candidate pool generated by the base model. Each row shows three frames capturing key stages of motion in the video.}
    \label{fig:comp_reward}
\end{figure}

\paragraph{Scorer comparison}
% \TODO{More VLM results in Tab2 and analysis}
~\Cref{fig:comp_reward} compares videos selected by different scorers from the same base-model candidate pool for each prompt. In the left case, additional objects appear in the videos selected by Qwen3-VL and WMReward, and the ball also disappears in the video selected by WMReward. By contrast, \rewardname{} selects a video in which the ball remains visible and rolls toward the duck. In the right case, both Qwen3-VL and WMReward select videos in which the ball flies upward unnaturally. By contrast, \rewardname{} selects a video in which the ball lands on the sand and remains at rest.
More qualitative results are provided in Appendix~\ref{app:qualitative_results}. 

% \input{resources/tables/physics_iq_benchmark_bon}

% \TODO{zheng, 9.23}

\subsection{Ablation Studies}

% \paragraph{Sampler components}
Table~\ref{tab:component_ablation} evaluates the contributions of latent refinement and physics guidance. 
Removing either component lowers the overall Physics-IQ score and VideoPhy PC.
Guidance alone performs similarly to the base sampler, whereas adding it to latent refinement raises the overall score. These results support the complementary roles of refinement and guidance, demonstrating the effectiveness of our sampling mechanism.
% \TODO{refine this, the two benchmarks do not directly reflect the two risks respectively?: consistent with our joint treatment of the two risks.}
Furthermore, our sensitivity analysis shows that \rewardname{} maintains stable performance across the tested hyperparameter settings; see Appendix~\ref{app:reward_validation} for details.

% For \rewardname{}, we evaluate the sensitivity of its hyperparameters and the results show that \rewardname{} is not sensitive to the hyperparameters, details can be found in Appendix~\ref{app:reward_validation}.
% \TODO{}

\begin{table*}[t]
\centering
\small
\setlength{\tabcolsep}{5pt}
\renewcommand{\arraystretch}{1.15}

\begin{adjustbox}{max width=\textwidth}
\begin{tabular}{@{}l cccc c c@{}}
\toprule
\multirow{3}{*}{\textbf{Setting}}
& \multicolumn{4}{c}{\textbf{Physics-IQ Metrics}}
& \multirow{3}{*}{\makecell{\textbf{Physics-IQ}\\\textbf{Score} $\uparrow$}}
& \multirow{3}{*}
{\makecell{\textbf{VideoPhy}\\\textbf{PC} $\uparrow$}} \\
\cmidrule(lr){2-5}
& \makecell{\textbf{Spatial}\\\textbf{IoU} $\uparrow$}
& \makecell{\textbf{Spatiotemporal}\\\textbf{IoU} $\uparrow$}
& \makecell{\textbf{Weighted}\\\textbf{Spatial IoU} $\uparrow$}
& \makecell{\textbf{MSE}\\\textbf{Score} $\uparrow$}
& 
& \\
\midrule

w/o both
& 0.3295 & \textbf{0.2442} & 0.1727
& \textbf{0.3074} & 26.34 & \underline{0.26} \\

w/ Latent Refinement
& \underline{0.3796} & 0.2322 & \underline{0.2109}
& \underline{0.2916} & \underline{27.86} & 0.24 \\

w/ Physics Guidance
& 0.3252 & \underline{0.2435} & 0.1680
& \textbf{0.3074} & 26.10 & \underline{0.26} \\

\rowcolor{gray!10}
\textbf{\samplername{}} (ours)
& \textbf{0.3927} & 0.2360 & \textbf{0.2243}
& 0.2846 & \textbf{28.44} & \textbf{0.30} \\

\bottomrule
\end{tabular}
\end{adjustbox}

\caption{\textbf{Component ablation of \samplername.}
We evaluate latent refinement and physics guidance individually
and together; w/o both uses neither component.
All settings use the same backbone, prompts, initial frames,
and paired seeds, without candidate selection.
The best and second-best results in each column are bold
and underlined, respectively.}
\label{tab:component_ablation}
\end{table*}

% \TODO{zheng, 9.23}

% \section{Further Anaylysis}

\section{Discussion}
Our approach may improve physical plausibility at the expense of generation diversity.
\samplername{} and \rewardname{} favor high-density regions of the learned distribution, potentially suppressing rare but physically valid outputs.
This trade-off may limit applications that require diverse video generation.
% or broad coverage in synthetic training data.
Preserving the diversity of valid outputs while reducing physical hallucinations remains an important direction for future work.
Furthermore, we discuss potential applications in synthetic-data generation for robotics and training guidance in Appendix~\ref{sec:applications}.
% We also discuss the potential applications of our method, including xxx, see details in Appendix~\ref{subsec:applications}.

\section{Conclusion}

We proposed a unified framework that distinguishes low-density sampling risk from high-density drifting risk in video generation.
Our analysis explains the complementary roles and limitations of model-internal and external signals: internal signals favor higher-likelihood generations, while external physical guidance helps address physical errors that the model itself assigns high likelihood.
Based on this framework, we proposed \samplername{} for video sampling and \rewardname{} for likelihood-based video evaluation.
Experiments demonstrate improvements in generating and selecting physically plausible videos. 
Beyond inference-time sampling, these risk measures could help select synthetic training videos and assess video models during post-training. 

\bibliography{iclr2027_conference}
\bibliographystyle{iclr2027_conference}

\clearpage
\appendix
% \section{Appendix}

\section{Mathematical Analysis}
\label{app:risk_mechanisms}

This section analyzes latent refinement and gradient guidance within our two-layer risk framework.
We interpret latent refinement through an idealized low-temperature Langevin process and gradient guidance through distribution reweighting based on the external loss.
These analyzes motivate combining the two mechanisms in \samplername{} to address low-density sampling risk and high-density drifting risk.

\paragraph{(i) Latent Refinement} 
At a fixed noise level $t$, latent refinement predicts a clean endpoint and adds fresh Gaussian noise to return it to the same noise level. It repeats this prediction--perturbation cycle before advancing to the next denoising step~\citep{jang2026selfrefining}. Under a Bayes-optimal endpoint predictor for the Gaussian noising process, Tweedie's formula~\citep{efron2011tweedie} gives
\begin{equation}
\mathbf{z}_t^{k+1} = \mathbf{z}_t^k + \sigma_t^2 \nabla_{\mathbf{z}_t} \log p_{\theta,t}(\mathbf{z}_t^k \mid \mathbf{c}) + \sigma_t \boldsymbol{\epsilon}_{k+1}, \qquad \boldsymbol{\epsilon}_{k+1} \sim \mathcal{N}(\mathbf{0}, \mathbf{I}),
\end{equation}
where $k$ indexes refinement iterations. The transition combines a step along the model score with a Gaussian perturbation. Its continuous Langevin counterpart has temperature $T = 1/2$ and stationary density
\begin{equation}
\pi_t(\mathbf{z}_t \mid \mathbf{c}) = \frac{p_{\theta,t}(\mathbf{z}_t \mid \mathbf{c})^2}{Z_t(\mathbf{c})},
\end{equation}
where $Z_t(\mathbf{c})$ is the normalization constant. Squaring the density increases the relative weight of higher-density regions, and this distribution satisfies
\begin{equation}
\mathbb{E}_{\mathbf{z}_t \sim \pi_t(\cdot \mid \mathbf{c})} [-\log p_{\theta,t}(\mathbf{z}_t \mid \mathbf{c})] \leq \mathbb{E}_{\mathbf{z}_t \sim p_{\theta,t}(\cdot \mid \mathbf{c})} [-\log p_{\theta,t}(\mathbf{z}_t \mid \mathbf{c})].
\end{equation}
The expected negative log-density here is the noisy-marginal counterpart of the risk in Definition~\ref{def:low_density_risk} which explains how latent refinement can reduce low-density sampling risk at a fixed noise level.

\paragraph{(ii) Gradient Guidance}
Gradient guidance predicts a clean endpoint during denoising, decodes it into a video, and evaluates the drifting risk $r_\omega$ through the external loss. The resulting gradient adjusts the sampling update toward outcomes that the evaluator considers more physically plausible~\citep{yuan2026wmreward}. The ideal reweighted target associated with this risk is
\begin{equation}
p_{\theta,\lambda}(\mathbf{z}_0 \mid \mathbf{c}) = \frac{p_\theta(\mathbf{z}_0 \mid \mathbf{c}) \exp[-\lambda r_\omega(\mathbf{z}_0 \mid \mathbf{c})]}{Z_\lambda(\mathbf{c})},
\end{equation}
where $\lambda \geq 0$ controls guidance strength and $Z_\lambda(\mathbf{c})$ is the normalizing constant. The exponential factor increases the relative weight of samples with lower drifting risk. Let $p_{\theta,\lambda,t}$ denote its marginal under the same noising process. The corresponding score is approximated as
\begin{equation}
\nabla_{\mathbf{z}_t} \log p_{\theta,\lambda,t}(\mathbf{z}_t \mid \mathbf{c}) \approx \nabla_{\mathbf{z}_t} \log p_{\theta,t}(\mathbf{z}_t \mid \mathbf{c}) - \lambda \nabla_{\mathbf{z}_t} r_\omega(\hat{\mathbf{z}}_{0\mid t} \mid \mathbf{c}).
\end{equation}
Prior work has reported limited gains in physical plausibility from gradient guidance alone in video generation~\citep{yuan2026wmreward}. Our analysis suggests that guidance can be more effective when low-density sampling risk is reduced or kept under control. 
% We therefore apply gradient guidance within latent refinement to reduce high-density drifting risk while controlling low-density sampling risk.

% \noindent\textbf{(ii) Tangential Amplifying Guidance.}
% TAG amplifies the tangential component of the Euler update:
% \begin{equation}
%     \Delta_t^{\mathrm{TAG}} = \left(\mathbf{P}_t^{\parallel} + \eta \mathbf{P}_t^{\perp}\right)\Delta_t^{\mathrm{E}}, \qquad \eta \geq 1,
% \end{equation}
% where $\mathbf{P}_t^{\parallel}$ and $\mathbf{P}_t^{\perp}$ are the radial and tangential projectors. Let $\mathbf{z}_t^{\mathrm{TAG}} = \mathbf{z}_t + \Delta_t^{\mathrm{TAG}}$ and $\mathbf{z}_t^{\mathrm{E}} = \mathbf{z}_t + \Delta_t^{\mathrm{E}}$ denote the paired states obtained from the same $\mathbf{z}_t$. With $\mathbf{s}_{\theta,t} = \nabla_{\mathbf{z}_t}\log p_{\theta,t}(\mathbf{z}_t \mid \mathbf{c})$, the first-order likelihood gain gives
% \begin{equation}
%     \mathbb{E}\left[-\log p_{\theta,t}(\mathbf{z}_t^{\mathrm{TAG}} \mid \mathbf{c})\right] - \mathbb{E}\left[-\log p_{\theta,t}(\mathbf{z}_t^{\mathrm{E}} \mid \mathbf{c})\right] \approx -(\eta - 1)b_t \mathbb{E}\left[\left\|\mathbf{P}_t^{\perp}\mathbf{s}_{\theta,t}\right\|_2^2\right],
% \end{equation}
% where $b_t > 0$ is the score coefficient in the ideal Euler update $\Delta_t^{\mathrm{E}} = a_t \mathbf{z}_t + b_t \mathbf{s}_{\theta,t}$. The leading term is non-positive, giving a first-order comparison of the noisy-marginal risk with Euler in the sense of Definition~\ref{def:low_density_risk}.

\section{Algorithms}
\label{app:algorithms}
Algorithm~\ref{alg:guided_latent_refine} presents one refinement iteration of \samplername{}, where gradient guidance updates the perturbation noise before the model re-predicts the clean endpoint.
For Wan2.2-TI2V-5B, we use 20 denoising steps, with three refinement iterations at steps 1--2 and one at steps 3--6.
For SkyReels-V2-I2V-1.3B, we use 50 denoising steps, with three refinement iterations at steps 3--7 and one at steps 8--17.
Algorithm~\ref{alg:sparse_cnf} details \rewardname{}, which estimates the negative log-likelihood per latent dimension using Heun integration and finite-difference divergence estimates.

\begin{algorithm}[H]
\caption{\samplername{}}
\label{alg:guided_latent_refine}
\renewcommand{\algorithmicrequire}{\textbf{Input:}}
\begin{algorithmic}[1]
\Require Predicted clean endpoint $\hat{\mathbf{z}}_{0\mid t}$, noise level $t$, guidance step size $\eta_t$, velocity field $v_\theta$, VAE decoder $\mathcal{G}$, external loss $\ell_\omega$, condition $\mathbf{c}$.
\State \textbf{Sample} $\boldsymbol{\epsilon} \sim \mathcal{N}(\mathbf{0}, \mathbf{I})$
\State \hspace{\algorithmicindent} $\mathbf{u}_t \leftarrow (1 - t) \hat{\mathbf{z}}_{0\mid t} - t \boldsymbol{\epsilon}$
\State \hspace{\algorithmicindent} $\mathbf{v} \leftarrow v_\theta(\mathbf{u}_t, t, \mathbf{c})$
\State \hspace{\algorithmicindent} $\hat{\mathbf{u}}_{0\mid t} \leftarrow \mathbf{u}_t - t \mathbf{v}$ \Comment{Probe endpoint}
\State \hspace{\algorithmicindent} $\mathcal{L} \leftarrow \ell_\omega\bigl(\mathcal{G}(\hat{\mathbf{u}}_{0\mid t}), \mathbf{c}\bigr)$
\State \hspace{\algorithmicindent} $\mathbf{g} \leftarrow \nabla_{\boldsymbol{\epsilon}} \mathcal{L}$ \Comment{Noise gradient}
\State \hspace{\algorithmicindent} $\mathbf{g} \leftarrow \operatorname{RMSNormalize}(\mathbf{g})$
\State \hspace{\algorithmicindent} $\boldsymbol{\epsilon}^{+} \leftarrow \boldsymbol{\epsilon} - \eta_t \mathbf{g}$ \Comment{Guided noise}
\State \hspace{\algorithmicindent} $\mathbf{u}_t^{+} \leftarrow (1 - t) \hat{\mathbf{z}}_{0\mid t} - t \boldsymbol{\epsilon}^{+}$
\State \hspace{\algorithmicindent} $\mathbf{v}^{+} \leftarrow v_\theta(\mathbf{u}_t^{+}, t, \mathbf{c})$ \Comment{Re-predict}
\State \hspace{\algorithmicindent} $\hat{\mathbf{z}}_{0\mid t}^{+} \leftarrow \mathbf{u}_t^{+} - t \mathbf{v}^{+}$ \Comment{Refined endpoint}
\State \Return $\hat{\mathbf{z}}_{0\mid t}^{+}, \mathbf{u}_t^{+}, \mathbf{v}^{+}$
\end{algorithmic}
\end{algorithm}

\begin{algorithm}[htbp]
\caption{\rewardname{}}
\label{alg:sparse_cnf}
\renewcommand{\algorithmicrequire}{\textbf{Input:}}
\begin{algorithmic}[1]
\Require Clean endpoint $\mathbf{z}_0$, base velocity field $v_\theta$, condition $\mathbf{c}$, time grid $0 = t_0 < \cdots < t_H = 1$, probe count $P$, relative perturbation scale $\delta_{\mathrm{rel}}$.
\State Sample independent probes $\mathbf{r}_1,\dots,\mathbf{r}_P$ \Comment{Fixed across steps}
\State $\mathbf{z}_{t_0} \leftarrow \mathbf{z}_0$, $A \leftarrow 0$, $D \leftarrow \dim(\mathbf{z}_0)$
\For{$j = 0,\dots,H-1$}
    \State $h_j \leftarrow t_{j+1} - t_j$
    \State $(\mathbf{v}_j, \hat{d}_j) \leftarrow \Call{Evaluate}{\mathbf{z}_{t_j}, t_j}$
    \State $\mathbf{z}_{\mathrm{E}} \leftarrow \mathbf{z}_{t_j} + h_j \mathbf{v}_j$ \Comment{Euler predictor}
    \State $(\mathbf{v}_j^{\mathrm{E}}, \hat{d}_j^{\mathrm{E}}) \leftarrow \Call{Evaluate}{\mathbf{z}_{\mathrm{E}}, t_{j+1}}$
    \State $A \leftarrow A + \frac{h_j}{2}(\hat{d}_j + \hat{d}_j^{\mathrm{E}})$ \Comment{Divergence integral}
    \State $\mathbf{z}_{t_{j+1}} \leftarrow \mathbf{z}_{t_j} + \frac{h_j}{2}(\mathbf{v}_j + \mathbf{v}_j^{\mathrm{E}})$ \Comment{Heun update}
\EndFor
\State $\hat{r}_\theta \leftarrow \frac{\|\mathbf{z}_{t_H}\|_2^2}{2} + \frac{D}{2}\log(2\pi) - A$ \Comment{Estimated NLL}
\State \Return $\hat{r}_\theta / D$ \Comment{NLL per dimension}
\Statex
\Function{Evaluate}{$\mathbf{z}, t$}
    \State $\mathbf{v} \leftarrow v_\theta(\mathbf{z}, t, \mathbf{c})$
    \State $\delta \leftarrow \delta_{\mathrm{rel}} \max\bigl(\|\mathbf{z}\|_2 / \sqrt{D},\, 0.1\bigr)$
    \State $\hat{d} \leftarrow \frac{1}{P\delta} \sum_{p=1}^{P} \mathbf{r}_p^\top \bigl[v_\theta(\mathbf{z} + \delta \mathbf{r}_p, t, \mathbf{c}) - \mathbf{v}\bigr]$ \Comment{Forward FD}
    \State \Return $\mathbf{v}, \hat{d}$
\EndFunction
\end{algorithmic}
\end{algorithm}

\section{Experimental Details}
\label{app:baselines-benchmarks}

We describe the baselines, benchmarks, and evaluation metrics used in our experiments. Sampling methods and best-of-$N$ video selection are evaluated separately.

\subsection{Baselines}
\label{app:baselines}

\subsubsection{Sampler Baselines}

\paragraph{Base model}
The pretrained video generator uses the reference Euler sampler with the official classifier-free guidance (CFG) settings, without additional latent refinement or physics guidance.

\paragraph{P\&P}~\citep{jang2026selfrefining}
P\&P interprets flow-matching video generators as denoising autoencoders and repeatedly predicts and perturbs latent states at a fixed noise level to favor stable model modes.

\paragraph{WMReward ($\nabla$)}~\citep{yuan2026wmreward}
We use its gradient-guidance variant, which guides sampling using prediction surprise from a pretrained latent world model as the external loss.

\paragraph{DreamWorld}~\citep{tan2026dreamworld}
DreamWorld jointly models video latents with motion, spatial, and semantic features extracted by pretrained models.
It uses constraint annealing during training and multi-source inner-guidance at inference.

\paragraph{VPT}~\citep{zheng2026enhancing}
VPT fine-tunes video generators with optical flow and entity-role maps, using modality-decoupled denoising and auxiliary loss annealing.
It further applies cross-step auto-guidance at inference to improve physical consistency.

\subsubsection{Scorer Baselines}

\paragraph{WMReward}~\citep{yuan2026wmreward}
We use prediction surprise from its pretrained latent world model to rank candidate videos, selecting the candidate with the lowest surprise.

\paragraph{GeoPhys}~\citep{interno2026geophys}
GeoPhys evaluates physical plausibility from the geometry of temporal feature trajectories extracted by frozen image encoders.
We reimplement its scoring method with the recommended DINOv3 backbone.

\paragraph{VLM scorers}
We evaluate Qwen3-VL-8B-Instruct~\citep{bai2025qwen3vl}, Gemma-4-E4B-it~\citep{gemmateam2026gemma4}, and Qwen3.5-9B~\citep{qwenteam2026qwen35}.
All three use the same binary answer-likelihood scoring procedure described in Appendix~\ref{app:vlm_results}.

For video selection, all scorers rank the same candidate pool for each condition, with random selection serving as the reference.

\subsection{Metrics}

\paragraph{Spatial IoU}
Compares temporally aggregated motion masks to measure whether motion occurs in the correct spatial regions.

\paragraph{Spatiotemporal IoU}
Compares frame-wise motion masks to assess both spatial and temporal alignment.

\paragraph{Weighted Spatial IoU}
Compares weighted motion maps to assess the location and magnitude of motion.

\paragraph{Mean Squared Error (MSE)}
Measures pixel-level differences between generated and reference continuations.

\subsection{Benchmarks}
\label{app:metrics}

\paragraph{Physics-IQ Verified}
We evaluate image-to-video generation using the Physics-IQ Verified metrics~\citep{radsch2026physicsiqverified}.
The benchmark compares generated continuations with recordings of real-world physical experiments.
Its four underlying metrics are described below.

We use the Verified aggregation, which normalizes each metric by the physical variation of the corresponding sample, clips each normalized subscore to $[0,1]$, and averages equally across metrics and samples.
The MSE score is transformed so that higher values indicate better performance.

\paragraph{VideoPhy}
We evaluate text-to-video generation on the 344-prompt VideoPhy test split~\citep{bansal2025videophy}.
We report semantic adherence (SA), which measures agreement with the text prompt, and physical commonsense (PC), which assesses whether the generated interactions follow physical laws.

\section{Details of Best-of-$N$ Selection}
\label{app:bon}

\subsection{VLM Scoring and Selection}
\label{app:vlm_results}

Following \citep{jang2025dreamgen}, we provide the VLM with a generated video and its input description, and ask it to assess physical plausibility with a binary response: \texttt{1} for physically plausible and \texttt{0} otherwise.
Following \citep{lin2024evaluating}, we use the log-probability of the positive answer token \texttt{1} as the score and select the highest-scoring candidate. The prompt used for VLM scoring is provided below.
\begin{quote}
\small\ttfamily
[Does the video show good physics dynamics and showcase a good alignment with the physical world for the prompt "{description}"? Please be a strict judge. If it breaks the laws of physics, please answer 0. Answer 0 for No or 1 for Yes. Reply only 0 or 1.]
\end{quote}

We use Qwen3-VL-8B-Instruct~\citep{bai2025qwen3vl} as the primary VLM.
We also evaluate Gemma-4-E4B-it~\citep{gemmateam2026gemma4} and Qwen3.5-9B~\citep{qwenteam2026qwen35}, using the same scoring procedure and candidate pools. ~\Cref{tab:vlm_bon_comparison} compares best-of-$N$ selection results across the three VLMs to examine how the evaluator choice affects the relative performance of the scorer methods. 
Across all three VLMs, the physical evaluation scores of the selected videos do not consistently improve as $N$ increases. This suggests that these VLMs still struggle to reliably identify physical errors and select more physically plausible videos.
\begin{table}[H]
\centering
\small
\setlength{\tabcolsep}{7pt}
\renewcommand{\arraystretch}{1.12}

\begin{adjustbox}{max width=\linewidth}
\begin{tabular}{@{}l c ccc@{}}
\toprule
\textbf{Sampler}
& \textbf{BoN}
& \textbf{Qwen3-VL}
& \textbf{Gemma-4}
& \textbf{Qwen3.5} \\
\midrule

\multirow{3}{*}{Base sampler}
& 2 & \textbf{27.73} & 26.48 & 27.27 \\
& 4 & \textbf{28.64} & 26.61 & 28.12 \\
& 8 & 28.18 & 26.55 & \textbf{29.42} \\

\midrule

\multirow{3}{*}{P\&P sampler}
& 2 & 28.33 & 27.08 & \textbf{28.69} \\
& 4 & 28.42 & 26.08 & \textbf{28.71} \\
& 8 & \textbf{28.74} & 26.24 & 28.22 \\

\midrule

\multirow{3}{*}{\samplername}
& 2 & 28.59 & 28.57 & \textbf{29.09} \\
& 4 & 28.16 & 26.66 & \textbf{29.21} \\
& 8 & 28.49 & 27.65 & \textbf{29.17} \\

\bottomrule
\end{tabular}
\end{adjustbox}

\caption{\textbf{Comparison of VLM-based selection across candidate pool sizes.}
We report Physics-IQ Verified scores for candidates selected
by different VLMs at $N \in \{2,4,8\}$.
Higher scores indicate better performance.
The highest score in each row is shown in bold.}
\label{tab:vlm_bon_comparison}
\end{table}

\subsection{Detailed Results}

\Cref{tab:physics_iq_bon_details} compare scorer methods with $N \in \{2,4,8\}$ candidates generated by the base sampler, P\&P, and \samplername{}.
\rewardname{} achieves the highest Physics-IQ Verified score in every setting, with scores increasing consistently across the tested candidate budgets for all three samplers.
Combined with \samplername{}, it achieves the best result in each budget.

\label{app:bon-details}
\begin{table}[H]
\centering
\small
\setlength{\tabcolsep}{7pt}
\renewcommand{\arraystretch}{1.12}

\begin{tabular}{@{}llccccc@{}}
\toprule
\textbf{Sampler}
& \textbf{BoN}
& \textbf{Random}
& \textbf{WMReward}
& \textbf{GeoPhys}
& \textbf{Qwen3-VL}
& \textbf{\rewardname{} (ours)} \\
\midrule

\multirow{3}{*}{Base sampler}
& 2 & 27.74 & \underline{28.84} & 26.57 & 27.73 & \textbf{28.99} \\
& 4 & 27.45 & \underline{30.75} & 26.50 & 28.64 & \textbf{30.77} \\
& 8 & 26.38 & \underline{30.51} & 25.67 & 28.18 & \textbf{30.88} \\

\midrule

\multirow{3}{*}{P\&P sampler}
& 2 & 27.28 & \underline{28.79} & 27.08 & 28.33 & \textbf{29.38} \\
& 4 & 27.09 & \underline{29.88} & 26.34 & 28.42 & \textbf{30.48} \\
& 8 & 27.02 & \underline{30.64} & 26.97 & 28.74 & \textbf{31.40} \\

\midrule

\multirow{3}{*}{\samplername{} (ours)}
& 2 & 27.54 & \underline{29.05} & 28.49 & 28.59 & \textbf{29.53} \\
& 4 & 27.04 & \underline{30.00} & 27.39 & 28.16 & \textbf{30.69} \\
& 8 & 27.18 & \underline{31.68} & 26.98 & 28.49 & \textbf{32.25} \\

\bottomrule
\end{tabular}

\caption{\textbf{Best-of-$N$ results with different candidate pool sizes.}
We report Physics-IQ Verified scores for selection methods using
$N \in \{2,4,8\}$ candidates from each sampler.
Random denotes uniform random selection.
Higher scores indicate better performance.
The best and second-best scores in each row are bold and underlined, respectively.}
\label{tab:physics_iq_bon_details}
\end{table}

\section{Hallucination Risk Across Samplers}
\label{app:risk_comparison}
We compare both risk measures across samplers on 100 cases in ~\Cref{tab:i2v_cnf_100_cases} and~\Cref{tab:i2v_high_density_drifting_risk}.
Low-density sampling risk is estimated by \rewardname{} on the final clean latents, while mean WMReward surprise is used as a proxy for high-density drifting risk.

\paragraph{Low-density sampling risk}
C-TAG, P\&P, CFG, and \samplername{} all reduce the estimated risk relative to Euler without CFG, whereas WMReward ($\nabla$) increases it.
Unlike WMReward ($\nabla$), \samplername{} reduces low-density sampling risk while applying gradient guidance.

% \begin{table}[t]
% \centering

% \small
% \setlength{\tabcolsep}{5pt}
% \renewcommand{\arraystretch}{1.12}

% \begin{tabular}{@{}l ccc cc@{}}
% \toprule
% \multirow{3}*{\textbf{Sampler}} & \multicolumn{3}{c}{\textbf{Risk Decomposition}}
% & & \\
% \cmidrule(lr){2-4}
% & \makecell{\textbf{Prior}\\\textbf{term}}
% & \makecell{\textbf{Volume}\\\textbf{term}}
% & \makecell{\textbf{Risk/dim}\\$\downarrow$}
% & \makecell{\textbf{$\Delta$Risk/dim}\\$\downarrow$}
% & \makecell{\textbf{Lower-risk}\\\textbf{cases}} \\
% \midrule
% Euler
% & 1.10503
% & -2.54120
% & -1.43617
% & 0.00000
% & -- \\

% C-TAG ($\lambda=1.25$)
% & 1.22623
% & -2.76701
% & -1.54079
% & -0.10462
% & 67/100 \\

% P\&P
% & 1.21343
% & -2.68283
% & -1.46940
% & -0.03323
% & 54/100 \\

% CFG
% & 1.21803
% & -2.70666
% & -1.48863
% & -0.05246
% & 60/100 \\
% \bottomrule
% \end{tabular}

% \caption{\textbf{Low-density sampling risk on 100 paired I2V cases.}
% The prior and volume terms sum to the CNF-estimated risk.
% All three quantities are reported per latent dimension.
% $\Delta$Risk is measured relative to the paired Euler baseline without
% classifier-free guidance; lower-risk cases count pairs with
% $\Delta$Risk $<0$.
% CNF estimates use $H=8$ Heun steps and $P=2$ shared Hutchinson probes.}
% \label{tab:i2v_cnf_100_cases}

% \end{table}

\begin{table}[H]
\centering
\small
\setlength{\tabcolsep}{5pt}
\renewcommand{\arraystretch}{1.12}

\begin{tabular}{@{}l cccc@{}}
\toprule
\multirow{2}{*}{\textbf{Sampler}}
& \multicolumn{3}{c}{\textbf{Risk Decomposition}}
& \multirow{2}{*}{\makecell{\textbf{$\Delta$Risk/dim}}} \\
\cmidrule(lr){2-4}
& \makecell{\textbf{Prior}\\\textbf{term}}
& \makecell{\textbf{Volume}\\\textbf{term}}
& \makecell{\textbf{Risk/dim}\\$\downarrow$}
& \\
\midrule
Euler
& 1.10503 & -2.54120 & -1.43617 & 0.00000 \\
C-TAG ($\lambda=1.25$)
& 1.22623 & -2.76701 & -1.54079 & -0.10462 \\
P\&P
& 1.21343 & -2.68283 & -1.46940 & -0.03323 \\
CFG
& 1.21803 & -2.70666 & -1.48863 & -0.05246 \\
WMReward ($\nabla$)
& 1.22506 & -2.21974 & -0.99468 & +0.44149 \\
\rowcolor{gray!10}
\samplername~(ours)
& 1.21706 & -2.72160 & -1.50454 & -0.06837 \\
\bottomrule
\end{tabular}

\caption{\textbf{Low-density sampling risk on 100 cases.}
The prior and volume terms sum to the low-density sampling risk.
All quantities are reported per latent dimension.
$\Delta$Risk is measured relative to the paired Euler baseline
without classifier-free guidance.
We use $H=8$ Heun steps and $P=2$ shared Hutchinson probes.
$\nabla$ denotes gradient guidance.}
\label{tab:i2v_cnf_100_cases}
\end{table}

\paragraph{High-density drifting risk}
Relative to the base sampler with CFG $=5$, P\&P slightly increases mean WMReward surprise, while WMReward ($\nabla$) and \samplername{} reduce it.
\samplername{} achieves the lowest mean risk among the evaluated methods.
Together, these results show that \samplername{} reduces both low-density sampling risk and high-density drifting risk.

\begin{table}[H]
\centering
\small
\setlength{\tabcolsep}{6pt}
\renewcommand{\arraystretch}{1.12}

\begin{tabular}{@{}l cc@{}}
\toprule
\textbf{Sampler}
& \textbf{Risk} $\downarrow$
& \textbf{$\Delta$Risk} \\
\midrule
Base (CFG $=5$)
& 0.414941 & 0.000000 \\
P\&P
& 0.415273 & +0.000332 \\
WMReward ($\nabla$)
& 0.414746 & -0.000195 \\
\rowcolor{gray!10}
\samplername~(ours)
& 0.414180 & -0.000761 \\
\bottomrule
\end{tabular}

\caption{\textbf{High-density drifting risk on 100 cases.}
We report mean WMReward surprise as a proxy for high-density drifting risk.
$\Delta$Risk is measured relative to the base sampler with CFG $=5$.
$\nabla$ denotes gradient guidance.}
\label{tab:i2v_high_density_drifting_risk}
\end{table}

\section{\rewardname{} Sensitivity}
\label{app:reward_validation}

\subsection{Sensitivity to $H$ and $p$}

% \paragraph{\rewardname~sensitivity analysis}
Table~\ref{tab:cnf_sensitivity} examines how the Heun step count $H$ and Hutchinson probe count $p$ affect \rewardname{}.
More steps or probes reduce the mean absolute error (MAE) relative to the reference NLL estimates, while rank correlations with the reference estimates remain close to one across the tested settings. These results indicate that the rankings produced by \rewardname{} are relatively insensitive to $H$ and $p$ within the tested range.
Table~\ref{tab:appendix_cnf_hp_ours} further shows that varying $H$ and $P$ within the tested range has little effect on candidate selection or physical evaluation results.

\begin{table*}[h]
\centering

\begin{subtable}[t]{0.48\textwidth}
\centering
\begin{adjustbox}{width=0.8\linewidth}
\begin{tabular}{c|c|cc}
\toprule
$H$ & \textbf{NFE} & \textbf{MAE} $\downarrow$ & $\rho$ $\uparrow$ \\
\midrule
\multicolumn{4}{c}{\textit{100 cases}} \\
\addlinespace[2pt]
4  & 24 & 0.067996 & 0.99600 \\
8  & 48 & 0.003014 & 0.99957 \\
16 & 96 & Ref.     & 1.00000 \\

\midrule
\multicolumn{4}{c}{\textit{25 case subset}} \\
\addlinespace[2pt]
16 & 96  & 0.001210 & 1.00000 \\
32 & 192 & Ref.     & 1.00000 \\
\bottomrule
\end{tabular}
\end{adjustbox}
\caption{\textbf{Heun integration steps.}}
\end{subtable}
\hfill
\begin{subtable}[t]{0.48\textwidth}
\centering
\begin{adjustbox}{width=0.8\linewidth}
\begin{tabular}{c|c|cc}
\toprule
$p$ & \textbf{NFE} & \textbf{MAE} $\downarrow$ & $\rho$ $\uparrow$ \\
\midrule
\multicolumn{4}{c}{\textit{100 cases}} \\
% \addlinespace[2pt]
1 & 32 & 0.000548 & 0.99994 \\
2 & 48 & 0.000322 & 0.99994 \\
4 & 80 & Ref.     & 1.00000 \\

\midrule
\multicolumn{4}{c}{\textit{25 case subset}} \\
% \addlinespace[2pt]
4 & 80  & 0.000256 & 1.00000 \\
8 & 144 & Ref.     & 1.00000 \\
\bottomrule
\end{tabular}
\end{adjustbox}
\caption{\textbf{Hutchinson probe count.}}
\end{subtable}

\caption{\textbf{Sensitivity of ~\rewardname{} to integration steps and probe count.}
We evaluate the full 100-case set and a fixed stratified 25-case subset.
We fix $p=2$ in (a), $H=8$ in (b), and $\delta_{\mathrm{rel}}=0.0125$ throughout.
$\rho$ denotes the Spearman rank correlation coefficient.
NFE counts model forward evaluations per case.} 
% \TODO{Add conclusion of these exps; add Ref. denotes}
\label{tab:cnf_sensitivity}

\end{table*}

\begin{table}[H]
\centering

\small
\setlength{\tabcolsep}{7pt}
\renewcommand{\arraystretch}{1.15}

\begin{adjustbox}{max width=\linewidth}
\begin{tabular}{@{}l cccc@{}}
\toprule

\multicolumn{5}{@{}l@{}}{
    \textbf{(a) Probe count} \quad ($H=8$)
} \\
\addlinespace[4pt]

\multirow{2}{*}{\textbf{Sampler}}
& \multicolumn{2}{c}{\textbf{Physics-IQ Score} $\uparrow$}
& \multirow{2}{*}{\textbf{$\Delta$Score}}
& \multirow{2}{*}{\textbf{Agreement}} \\
\cmidrule(lr){2-3}
& $P=1$ & $P=2$ & & \\
\midrule

Base sampler
& 30.88 & 30.88 & $+0.00$ & $198/198$ \\

P\&P sampler
& 31.28 & 31.40 & $+0.12$ & $197/198$ \\

\samplername
& 32.25 & 32.25 & $+0.00$ & $198/198$ \\

\midrule
\multicolumn{5}{@{}l@{}}{
    \textbf{(b) Integration steps} \quad ($P=1$)
} \\
\addlinespace[4pt]

\multirow{2}{*}{\textbf{Sampler}}
& \multicolumn{2}{c}{\textbf{Physics-IQ Score} $\uparrow$}
& \multirow{2}{*}{\textbf{$\Delta$Score}}
& \multirow{2}{*}{\textbf{Agreement}} \\
\cmidrule(lr){2-3}
& $H=4$ & $H=8$ & & \\
\midrule

\samplername
& 32.17 & 32.25 & $+0.08$ & $191/198$ \\

\bottomrule
\end{tabular}
\end{adjustbox}

\caption{\textbf{Effect of estimation settings on \rewardname.}
We compare Physics-IQ Verified scores using different numbers
of Heun steps ($H$) and Hutchinson probes ($P$).
$\Delta$Score is the change from the left setting to the right.
Agreement counts prompts with the same selected candidate.}
\label{tab:appendix_cnf_hp_ours}

\end{table}

\subsection{Sensitivity to $\delta_{\mathrm{rel}}$}
\label{app:delta_sensitivity}

Table~\ref{tab:fd_step_size} examines the sensitivity of \rewardname{} to $\delta_{\mathrm{rel}}$ by comparing forward finite-difference estimates with VJP-based estimates.
The default setting produces rankings close to the VJP reference, while the smallest and largest tested perturbation scales lead to larger estimation errors.

% Required packages: booktabs, makecell, adjustbox

\begin{table}[H]
\centering
\small
\setlength{\tabcolsep}{6pt}
\renewcommand{\arraystretch}{1.15}

\begin{adjustbox}{max width=\textwidth}
\begin{tabular}{@{}l ccccccc@{}}
\toprule
\makecell[l]{\textbf{Estimator /}\\$\delta_{\mathrm{rel}}$}
& \makecell{\textbf{Mean}\\\textbf{Risk/dim}}
& \makecell{\textbf{Bias}\\\textbf{vs. VJP}}
& \textbf{MAE} $\downarrow$
& \textbf{Pearson} $\uparrow$
& \textbf{Spearman} $\uparrow$
& \textbf{Kendall} $\uparrow$
& \makecell{\textbf{Mean rank}\\\textbf{shift} $\downarrow$} \\
\midrule

VJP
& -1.48701 & -- & -- & -- & -- & -- & -- \\

\midrule
\multicolumn{8}{@{}l@{}}{\textit{Forward finite differences}} \\
\addlinespace[3pt]

0.0050
& -1.51387 & -0.02686 & 0.08914
& 0.95765 & 0.93638 & 0.80889 & 7.30 \\

0.0100
& -1.47179 & +0.01523 & 0.02618
& 0.99493 & 0.99123 & 0.93374 & 2.52 \\

0.0125
& -1.46672 & +0.02030 & 0.02890
& 0.99456 & 0.99191 & 0.93051 & 2.70 \\

0.0250
& -1.47240 & +0.01461 & 0.01739
& 0.99846 & 0.99735 & 0.96485 & 1.40 \\

0.0500
& -1.40883 & +0.07818 & 0.07818
& 0.99800 & 0.99740 & 0.96687 & 1.40 \\

0.1000
& -1.17969 & +0.30732 & 0.30732
& 0.97516 & 0.97786 & 0.87475 & 4.54 \\

\bottomrule
\end{tabular}
\end{adjustbox}

\caption{\textbf{Finite-difference accuracy across perturbation scales.}
We compare forward finite differences with VJP-based estimates.
$\delta_{\mathrm{rel}}$ denotes the relative perturbation scale.
Bias and MAE are computed relative to VJP, while correlations
and mean rank shift measure agreement across evaluated samples.}
\label{tab:fd_step_size}
\end{table}

% \subsection{Effects on Selection}
% \label{app:hp_sensitivity}
% \input{resources/tables/appendix/cnf_paras}

% \subsection{}
% \input{resources/tables/appendix/ref-model-physicsiq}

\section{More Qualitative Results}
\label{app:qualitative_results}

\subsection{Qualitative Results on Physics-IQ}
\begin{figure}[p]
    \centering
    \includegraphics[width=\linewidth]
    {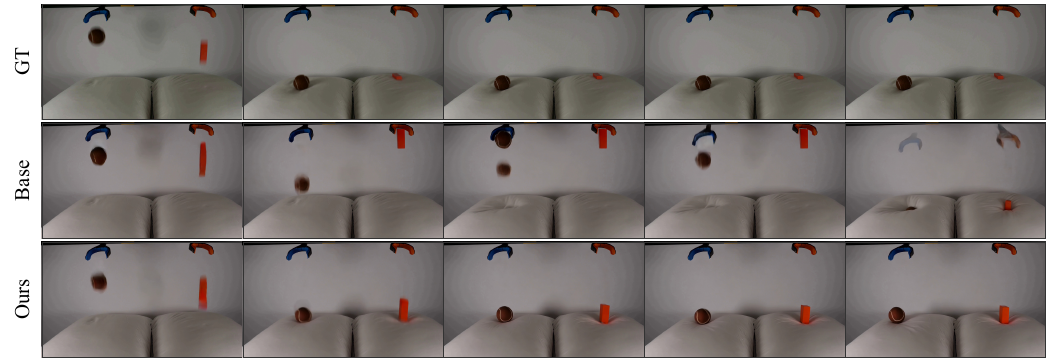}
    \par\vspace{3pt}
    \includegraphics[width=\linewidth]
    {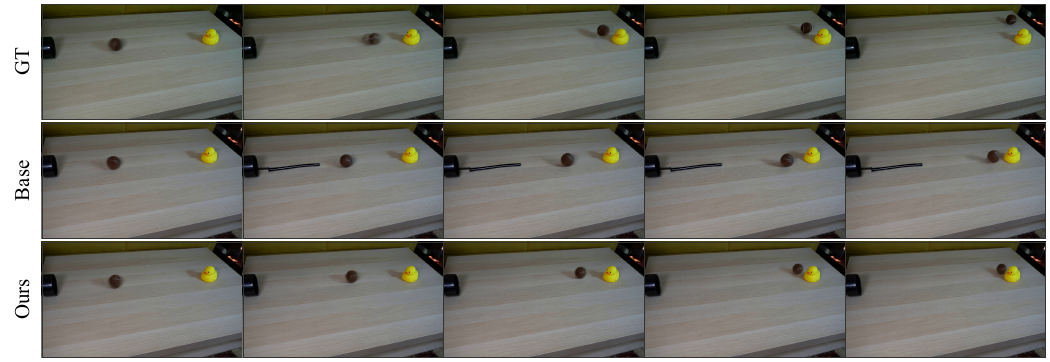}
    \par\vspace{3pt}
    \includegraphics[width=\linewidth]
    {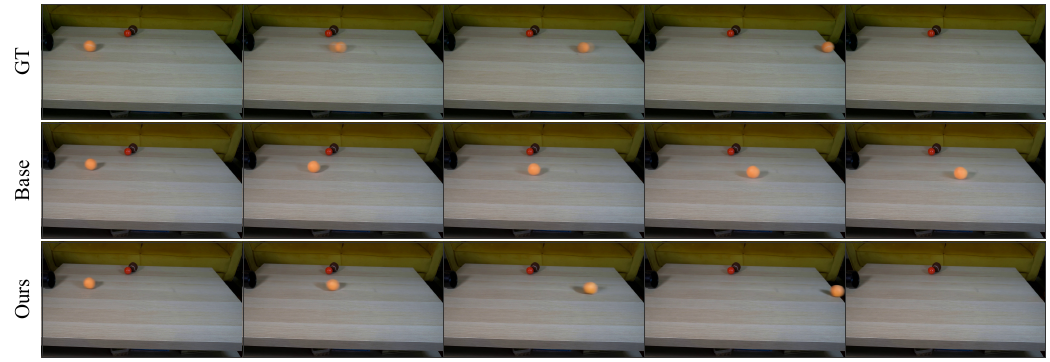}
    \par\vspace{3pt}
    \includegraphics[width=\linewidth]
    {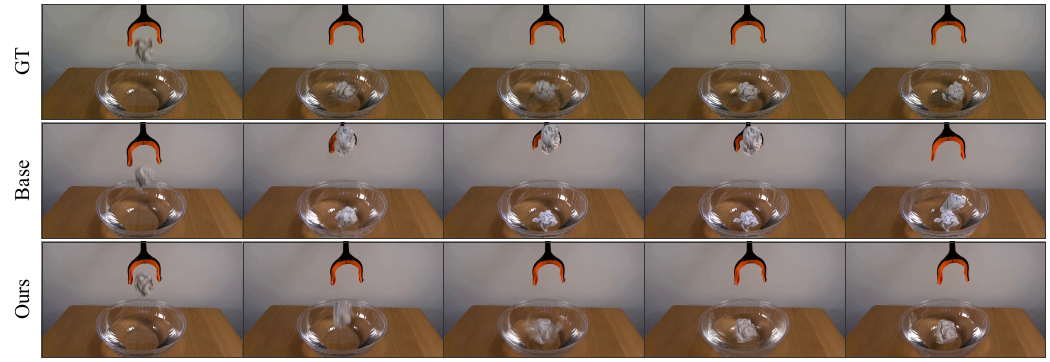}

    \caption{
        \textbf{Qualitative comparisons on Physics-IQ.}
        Each example shows GT, Base, and Ours from top to bottom,
        with five frames arranged chronologically.
        Ours denotes videos generated with \samplername{} and selected using \rewardname{}.
    }
    \label{fig:physics_iq_comparison_1}
\end{figure}

\begin{figure}[p]
    \centering
    \includegraphics[width=\linewidth]
    {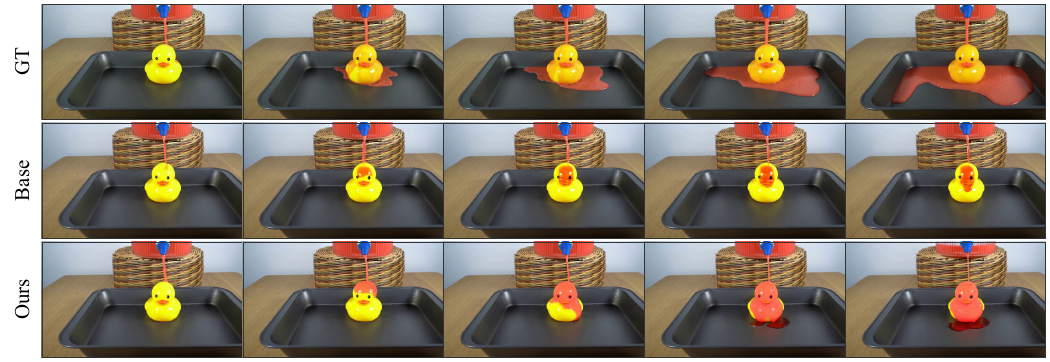}
    \par\vspace{3pt}
    \includegraphics[width=\linewidth]
    {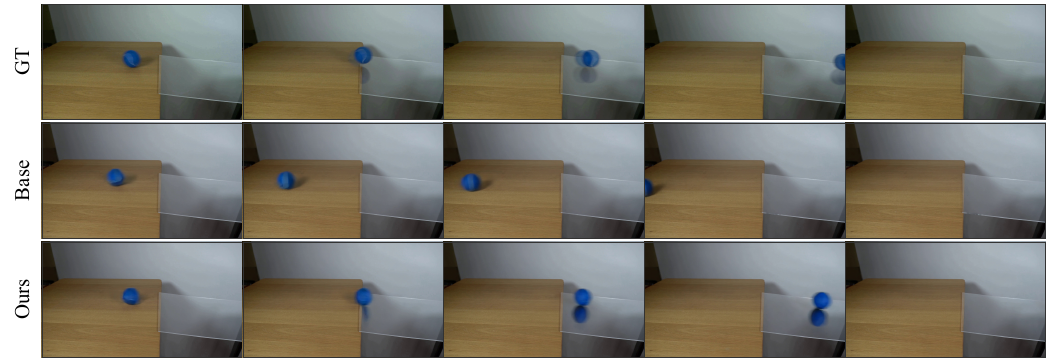}
    \par\vspace{3pt}
    \includegraphics[width=\linewidth]
    {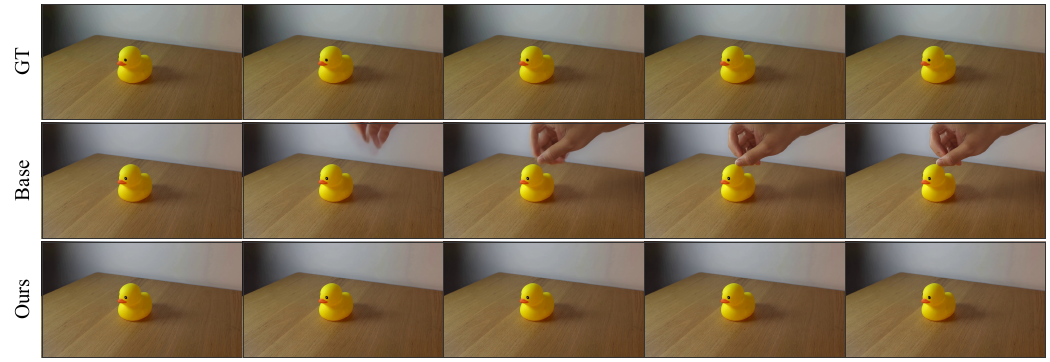}
    \par\vspace{3pt}
    \includegraphics[width=\linewidth]
    {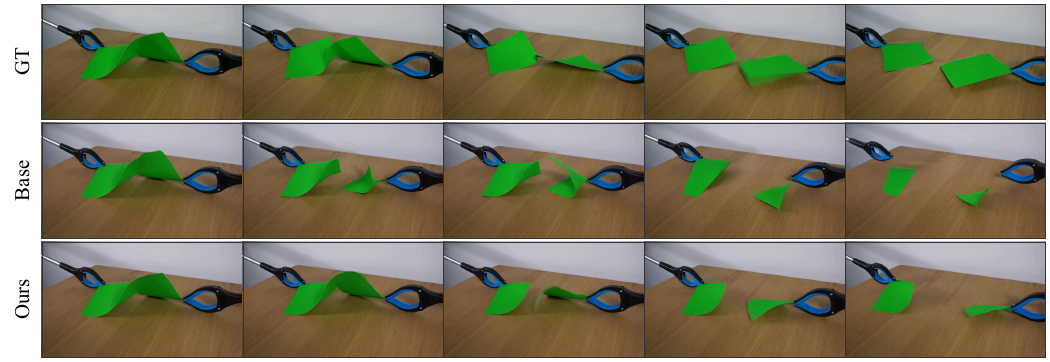}

    \caption{
        \textbf{Additional qualitative comparisons on Physics-IQ.}
        The layout follows Figure~\ref{fig:physics_iq_comparison_1}.
    }
    \label{fig:physics_iq_comparison_2}
\end{figure}

\Cref{fig:physics_iq_comparison_1} and~\Cref{fig:physics_iq_comparison_2} compare our method with the baselines on Physics-IQ, focusing on generated motion and object interactions. 
Compared with the baselines, our method produces more physically plausible motion and object interactions in these examples.
These observations are consistent with the quantitative results and support combining \samplername{} for generation with \rewardname{} for candidate selection.
% \TODO{add fig}

\subsection{Qualitative Results on VideoPhy}

\begin{figure}[h]
    \centering
    \begin{minipage}[t]{0.49\linewidth}
        \vspace{0pt}
        \centering
        {\footnotesize\itshape
        ``A brush sweeps blue paint across a canvas.''\par}
        \vspace{2pt}
        \includegraphics[width=\linewidth]
        {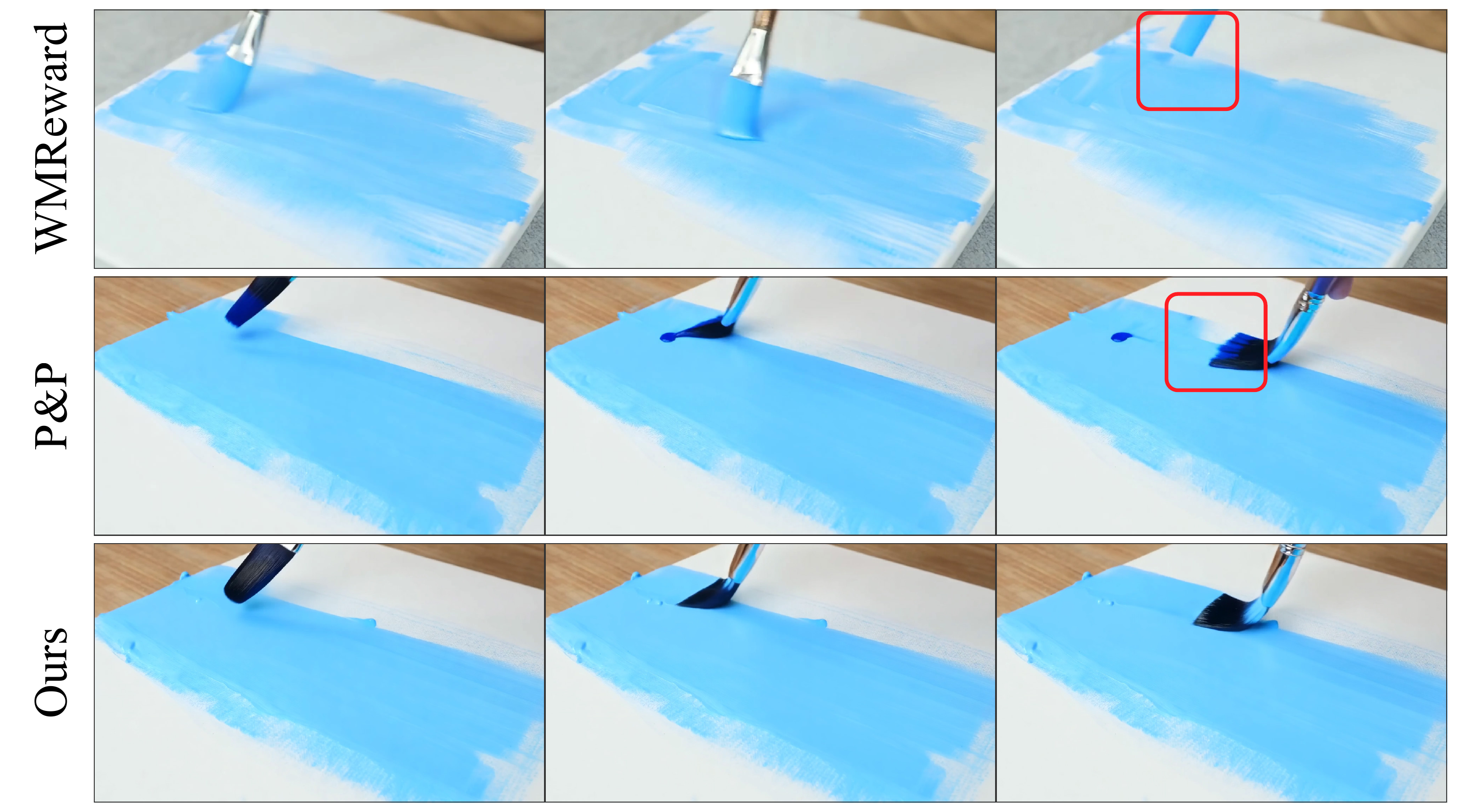}
    \end{minipage}\hfill
    \begin{minipage}[t]{0.49\linewidth}
        \vspace{0pt}
        \centering
        {\footnotesize\itshape
        ``Water flowing out from a pipe.''\par}
        \vspace{2pt}
        \includegraphics[width=\linewidth]
        {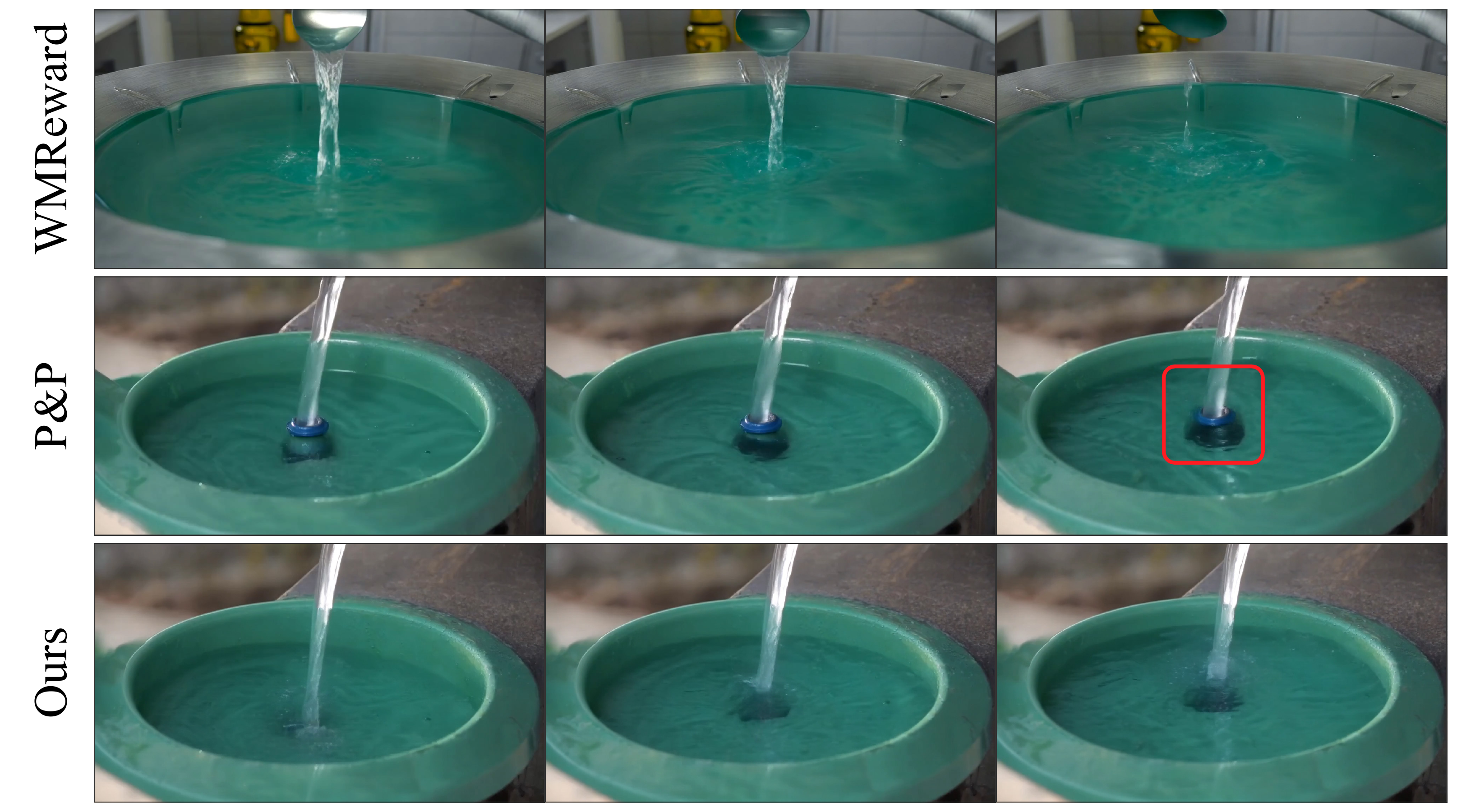}
    \end{minipage}

    \caption{\textbf{Qualitative comparison of T2V generation on VideoPhy.}
    For each prompt, we compare different samplers using three representative frames from each generated video. \samplername{} improves physical plausibility in both cases and helps correct inconsistencies in the generated interactions.}
    \label{fig:t2v_comp_sampling}
\end{figure}

\begin{figure}[p]
    \centering
    \includegraphics[width=\linewidth]
    {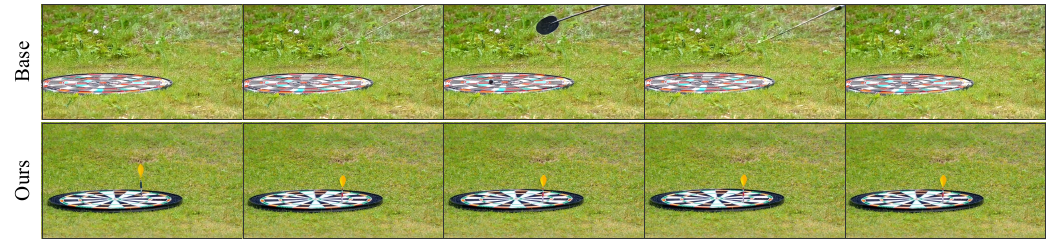}
    \par\vspace{3pt}
    \includegraphics[width=\linewidth]
    {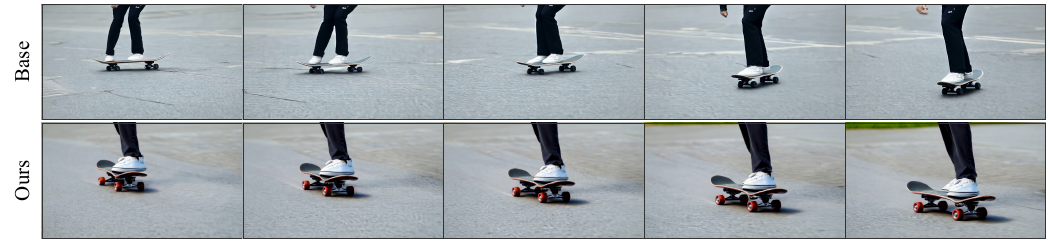}
    \par\vspace{3pt}
    \includegraphics[width=\linewidth]
    {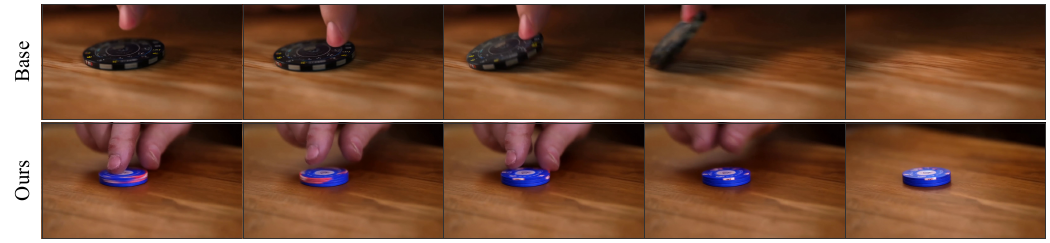}
    \par\vspace{3pt}
    \includegraphics[width=\linewidth]
    {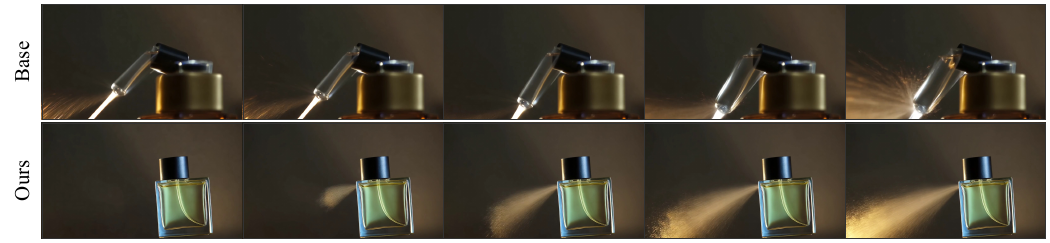}
    \par\vspace{3pt}
    \includegraphics[width=\linewidth]
    {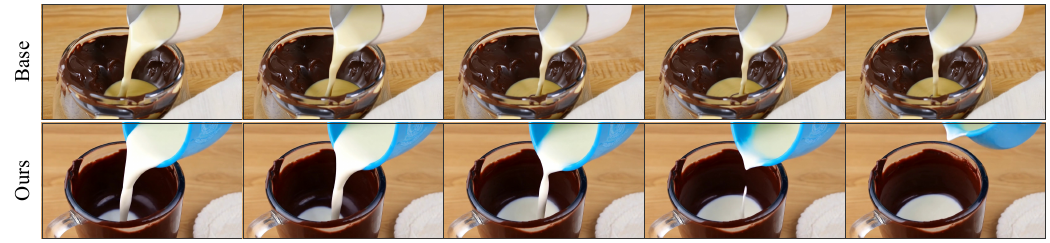}
    \par\vspace{3pt}
    \includegraphics[width=\linewidth]
    {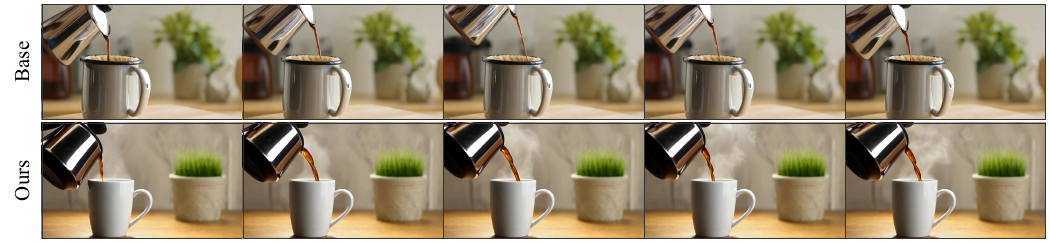}
    \caption{
        \textbf{Qualitative comparisons on VideoPhy.} For each prompt, we show five frames from videos produced by the base model and \samplername{} (Ours).
    }
    \label{fig:videophy_comparsion}
\end{figure}

\Cref{fig:t2v_comp_sampling} compares T2V results with P\&P and WMReward ($\nabla$).
In the left case, \samplername{} produces a cleaner contact between brush and canvas.
In the right case, it produces a more natural interaction with the water surface.
These results illustrate \samplername{} has improvements in physical details.
Additional qualitative comparisons on VideoPhy are provided in ~\Cref{fig:videophy_comparsion}.

% 正文：三个 case 竖向排列

% \clearpage

\section{Applications}
\label{sec:applications}

\paragraph{Reliable data generator}
\samplername{} supports more reliable synthetic-data generation for critical downstream tasks, such as robotics and autonomous driving.
\samplername{} improves the physical plausibility of generated motion and interactions, while \rewardname{} provides a physics-aware criterion for selecting candidate videos. 
Future work could extend our risk analysis to other rectified-flow-based generative tasks~\citep{esser2024scaling,liu2024instaflow,li2026adaptive} to guide the generation of more reliable synthetic data for downstream tasks~\citep{tian2024learning,trabucco2024effective}.

\paragraph{Training-free signal for training}
\rewardname~could also provide a diagnostic signal for monitoring video models during training without training an additional evaluator.
A possible protocol is to generate videos from successive checkpoints under fixed conditions and sampling settings, then score them using our \rewardname{}.
The resulting likelihood-based risk could help track hallucination-related failure patterns and guide checkpoint selection.

\end{document}